\documentclass[manuscript,screen]{acmart}

\usepackage{graphicx}
\usepackage{amsmath}
\usepackage{multirow}
\usepackage{algorithm}
\usepackage{algorithmicx}  
\usepackage{algpseudocode}
\usepackage{array}
\usepackage{xcolor}
\usepackage[caption=false]{subfig}
\usepackage{balance}

\graphicspath{{figures/}}

\floatname{algorithm}{Algorithm}

\algrenewcommand\algorithmicforall{\textbf{for each}}

\newcolumntype{P}[1]{>{\centering\arraybackslash}p{#1}}

\AtBeginDocument{%
  \providecommand\BibTeX{{%
    \normalfont B\kern-0.5em{\scshape i\kern-0.25em b}\kern-0.8em\TeX}}}

\setcopyright{acmcopyright}
\copyrightyear{2022}
\acmYear{2022}
\acmDOI{10.1145/1122445.1122333}

\acmJournal{TSAS}
\acmVolume{1}
\acmNumber{1}
\acmArticle{1}
\acmMonth{1}

\begin{document}

\title{Multiple-level Point Embedding for Solving Human Trajectory Imputation with Prediction}

\author{Kyle K. Qin}
\affiliation{%
  \institution{RMIT University}
  \streetaddress{124 La Trobe St}
  \city{Melbourne}
  \state{VIC}
  \country{Australia}
  \postcode{3000}
}
\email{kyle.qin@hotmail.com}

\author{Yongli Ren}
\affiliation{%
  \institution{RMIT University}
  \streetaddress{124 La Trobe St}
  \city{Melbourne}
  \state{VIC}
  \country{Australia}
  \postcode{3000}
}
\email{yongli.ren@rmit.edu.au}

\author{Wei Shao}
\affiliation{%
  \institution{RMIT University}
  \streetaddress{124 La Trobe St}
  \city{Melbourne}
  \state{VIC}
  \country{Australia}
  \postcode{3000}
}
\email{wei.shao@rmit.edu.au}

\author{Brennan Lake}
\affiliation{%
  \institution{Cuebiq Inc.}
  \streetaddress{15 West 27th Street}
  \city{New York}
  \state{NY}
  \country{USA}
  \postcode{10001}
}
\email{blake@cuebiq.com}

\author{Filippo Privitera}
\affiliation{%
  \institution{Cuebiq Inc.}
  \streetaddress{15 West 27th Street}
  \city{New York}
  \state{NY}
  \country{USA}
  \postcode{10001}
}
\email{fprivitera@cuebiq.com}

\author{Flora D. Salim}
\affiliation{%
  \institution{University of New South Wales}
  \streetaddress{School of Computer Science and Engineering, Engineering Rd}
  \city{Kensington}
  \state{NSW}
  \country{Australia}
  \postcode{2052}
}
\email{flora.salim@unsw.edu.au}

\renewcommand{\shortauthors}{Qin et al.}

\begin{abstract}
Sparsity is a common issue in many trajectory datasets, including human mobility data. This issue frequently brings more difficulty to relevant learning tasks, such as trajectory imputation and prediction. Nowadays, little existing work simultaneously deals with imputation and prediction on human trajectories. This work plans to explore whether the learning process of imputation and prediction could benefit from each other to achieve better outcomes. And the question will be answered by studying the coexistence patterns between missing points and observed ones in incomplete trajectories. More specifically, the proposed model develops an imputation component based on the self-attention mechanism to capture the coexistence patterns between observations and missing points among encoder-decoder layers. Meanwhile, a recurrent unit is integrated to extract the sequential embeddings from newly imputed sequences for predicting the following location. Furthermore, a new implementation called Imputation Cycle is introduced to enable gradual imputation with prediction enhancement at multiple levels, which helps to accelerate the speed of convergence. The experimental results on three different real-world mobility datasets show that the proposed approach has significant advantages over the competitive baselines across both imputation and prediction tasks in terms of accuracy and stability.
\end{abstract}



\maketitle

\section{Introduction}
Mining knowledge on datasets such as time series and human mobility data has received much attention from the public \cite{chen2016promises,fedus2018maskgan,liang2016online,teixeira2021impact,naghizade2020privacy}. However, missing values that commonly exist in this type of dataset can implicitly influence the quality of the analysis and learning process. As we know, data on human mobility at a fine-grained scale can help with understanding people’s movement patterns, activities, and potential intentions, allowing customized recommendations and services to be delivered to individuals or groups. However, human trajectory data is frequently incomplete in practice for many reasons, such as power or bandwidth limitations on sensor-based devices and varying communication frequency of signals between devices and servers. It has been mentioned that the sparsity of data could cause deterioration of performance on learning human activities or movements \cite{naghizade2018contextual,liu2019naomi}. Therefore, imputing missing values in a trajectory becomes a fundamental problem for other learning tasks, such as predictions.

Human mobility prediction or imputation has become prevalent, with various types of mobility data available for collection and processing. Previous works of human trajectory prediction mainly focused on forecasting future positions at different levels of granularity, including coordinates \cite{giuliari2020transformer,gupta2018social,sadri2018will}, specific locations such as Point-of-Interests (POIs) \cite{feng2018deepmove,wang2018exploiting,wang2020relation,zhou2020semi} and grids or regions on map \cite{gidofalvi2012and,monreale2009wherenext}. Intuitively, the imputation of human trajectories can be formulated as the problem of time series imputation, which a number of relevant approaches have solved. Previous algorithms proposed for sequence imputation generally leverage the advantages of Generative Adversarial Networks (GANs) \cite{fedus2018maskgan,luo2018multivariate,luo2019e2gan} or Recurrent Neural Networks (RNNs) \cite{cao2018brits}. However, those generative models that are autoregressive could be vulnerable to compounding errors in long-range temporal sequence modeling, and the recurrent models may face issues of vanishing gradients and difficult training. Recently, two non-autoregressive models \cite{liu2019naomi,qi2020imitative} were claimed to have more merits for imputing values in sequential data such as traffic flows and pedestrians' trajectories in a specific scene. Nonetheless, these approaches first lack explicit observation and evaluation of the imputation of human mobility in the scope of the city. As we know, compared with walking or running trajectories in a small scene, the patterns and factors could be dramatically varied when the distribution of movements is spread at the macro city scope (e.g., check-in data). We would encounter more challenges in this kind of dataset which is normally accompanied by arbitrary movements, sparsity, and no uniform frequency of points collection. Secondly, the specific dependencies or coexistence patterns between visited locations and missing ones in trajectories were not well considered during the training by the previous methods. Moreover, little existing work deals with prediction while simultaneously imputing values in the data of trajectories.

This paper plans to tackle the imputation issue by studying the coexistence patterns or dependencies between missing points and observed ones while incomplete human trajectories are given. Meanwhile, we argue that while solving the data sparsity issue, the imputation and prediction tasks can complement each other. In other words, we develop an approach that could achieve mutually beneficial effects between both tasks with interactive optimization. As we know, information will be lost when passing messages in a sequential manner of learning. Missing values are often randomly distributed in human movement trajectories, potentially inhibiting the efficacy of imputation in a  traditional sequential manner (e.g., RNNs). Transformer Networks \cite{vaswani2017attention} can effectively learn representations of observations for prediction purposes by weighting the values of sequences via the attention mechanism. This method is naturally suitable for learning in a non-sequential way. We propose a new approach called multIple-level poiNt embeddinG for solving tRajectory imputAtion wIth predictioN (INGRAIN). INGRAIN can effectively capture coexistence dependencies among observed and missing values via multi-head attention in encoder-decoder layers, which helps ameliorate the imputation efficiency. Meanwhile, the model effectively integrates with a recurrent unit to extract sequential embeddings from newly imputed sequences to predict the next moving positions. For flexible learning, a new process called the Imputation Cycle is designed to enable progressive imputation with prediction at multiple levels by constraining the number of missing points for each imputation recurrence. In addition, the model delivers messages from the imputation module to RNN units to generate sequential embeddings for forecasting purposes. In summary, the key contributions of this paper are as follows:
\begin{itemize}
    \item We propose one new framework that integrates both autoregressive and non-autoregressive components to impute missing points in human trajectories and predict future movements.
    \item A model is established for trajectory imputation with prediction based on gradual amelioration at multiple levels via setting the granularity of imputing points. It is applicable to different human mobility datasets.
    \item Comprehensive evaluations are conducted to show the efficiency and effectiveness of our model on three real-world human trajectory datasets. The paper provides insights into how the method satisfies accurate estimations on missing points and next positions and how trade-offs could be handled in this type of cooperative learning.
\end{itemize}

The rest of the paper is organized as follows: Section \ref{sec:relate_work} introduces the related work, and Section \ref{sec:problem_def} provides the definitions for the problem of human trajectory imputation with prediction. Section \ref{sec:proposed_work} presents the details of the proposed solution and its important components. Section \ref{sec:experiment} shows the experimental results, and the conclusion is given in Section \ref{sec:conclusion}.

\section{Related Work}
\label{sec:relate_work}
\subsection{Human Trajectory Prediction}
Many studies have been devoted to human trajectory prediction. They can be categorized according to the granularity of targeted locations in prediction. A part of the literature has focused on forecasting grids or regions for next movements on maps \cite{gidofalvi2012and,monreale2009wherenext}, whereas other parts have explored predicting coordinates \cite{giuliari2020transformer,gupta2018social,sadri2018will} or POIs \cite{feng2018deepmove,wang2018exploiting,wang2020relation} in the future. We can further classify the relevant literature from the perspective of traveling scope in human mobility data: some studies \cite{feng2018deepmove,sadri2018will,wang2018exploiting,wang2020relation} have concentrated on predicting locations of people across city areas that are often sparsely or scattered populated, and others \cite{alahi2016social,giuliari2020transformer,gupta2018social} have examined the motions of persons in smaller-scale scenes such as crowds on the street or customers on the floor of a building. The patterns and factors of human movement could be particularly distinct while the scale of traveling distance is changed. For example, by extracting grid-based stay time statistics of users and periodically analyzing their frequency of visiting regions \cite{gidofalvi2012and}, the Inhomogeneous Continuous-Time Markov model was used to capture temporal and sequential regularities for predicting the leaving time of objects in regions and their next locations. There is a certain positive correlation between people's morning trajectories and corresponding afternoon trajectories \cite{sadri2018will} in cities, so the integrated similarity metric was developed to estimate similar segments of trajectories with temporal segmentation and temporal correlation extraction. In a series of RNNs, LSTM \cite{hochreiter1997long} and Gated Recurrent Units (GRU) \cite{chung2014empirical} are two popular variants that effectively moderate short-term memory and can help to avoid the vanishing gradient problem. Social Long Short-Term Memory \cite{alahi2016social} was proposed based on LSTM to estimate the motions of pedestrians among the crowd in different scenes while taking into account the navigation of all their walking neighbors in a shared site.

\subsection{Missing Data Imputation}
Imputation of missing values on trajectory datasets has become an indispensable work in many applications. Previously, a considerable number of methods for trajectory completion focused on inferring missing portion of traffic trajectories from sparse GPS samples based on the geometry of road network \cite{yin2018feature,lou2009map,zheng2012reducing,li2016knowledge}. For instance, the History-based Route Inference System (HRIS) \citep{zheng2012reducing} was established with a set of new mapping algorithms that could effectively extract and learn the traveling patterns from historical trajectories and incorporate them into the route inference process. Along with estimating the traffic flows across junctions in a road network, Li et al. utilized the GPS samples within each flow cluster to achieve fine-level completion of individual trajectories \cite{li2016knowledge}. Yin et al. developed a map-matching algorithm by evaluating the traveling cost of candidate routes and considering the distance feature and road selection behavior of users \cite{yin2018feature}. However, the imputation methods for traffic trajectories on roads are hard to adapt to outdoor human mobility in city areas. Those road network mapping techniques could become ineffective while human beings' movements are relatively arbitrary on the map and affected by a variety of factors.

More recent research related to missing value imputation has mainly relied on techniques to impute sequential data including Matrix Factorisation \cite{naghizade2018contextual}, GANs \cite{fedus2018maskgan,luo2018multivariate,luo2019e2gan,yoon2018gain} or RNNs \cite{cao2018brits,yoon2018estimating}. Naghizade et al. \cite{naghizade2018contextual} proposed a contextual model to predict information at missing locations in sparse indoor trajectories via using Graph-regularised Non-negative Matrix Factorisation with consideration of implicit social ties among individuals. A model called BRITS \cite{cao2018brits} was built on RNNs to directly learn missing values for time series in a bidirectional recurrent dynamical system without strong assumptions. In GAIN \cite{yoon2018gain}, the generator imputes the missing values conditioned on observed data, and the discriminator then attempts to identify which parts of the conjectured vector are actually observed and which are imputed. The former module focuses on the imputation quality, and the latter is forced to learn according to real data distribution. In contrast, MASKGAN \citep{fedus2018maskgan} explicitly trains the generator to produce high-quality samples for infilling text on sentences. GRUI \cite{luo2018multivariate}, an autoregressive model based on GAN, was developed for the imputation of multivariate time series, such as electronic medical record datasets or air quality and weather data. Additionally, Liu et al. recently presented a non-autoregressive model NAOMI \cite{liu2019naomi} to impute missing values in different sequential datasets, such as traffic flows and trajectories of basketball players. The missing values are filled recursively from coarse to fine-grained resolutions via a forward and backward RNN-based model. NAOMI considers multiple-resolution imputation, wherein new imputations will be performed based on the previous inferences of values.

\subsection{Attention Mechanism}
In the learning process, recurrent models can make predictions by transiting successive dependencies of observations from the beginning of entire sequences. That gives recurrent models the expressive ability to deal with long sequential data while maintaining hidden states. Nowadays, autoregressive models such as the Transformer Network are competitive with or even supersede RNNs on a diverse set of missions. Transformer Network is claimed to effectively weigh and learn representations over available observations via the multi-head attention mechanism. The original Transformer \cite{vaswani2017attention} was established to model and predict sequences in the natural language processing field. Subsequently, GATs \cite{velivckovic2017graph} was developed to operate on graph-structured data, leveraging self-attentional layers to address weighting nodes for graph convolutions. Recently, it was adopted by Giuliari et al. \cite{giuliari2020transformer} to forecast the future motions of people in different scenes, which renders better performance than the LSTM-based and Linear approaches.

\subsection{Main Research Gaps}
In this paper, we attempt to examine whether the imputation of missing points in human historical mobility can bring sake to the prediction of future movements, which is rarely investigated by the existing works. As we mentioned, our objective focus on inferring missing locations of daily human mobility in city areas. The road network mapping techniques for vehicle traffic trajectories can not adapt well to such datasets as the distribution of human beings' movements is relatively arbitrary without strictly following the road networks. And human activities could be affected heavily by social connections, professions, and weather conditions. In addition, many GAN-based approaches, such as MASKGAN, GRUI, and GAIN were not designed or tested for complex human mobility data in terms of constructing coordinates of locations with Spatio-temporal dimensions. Recently, NAOMI and another variant called SingleRes \cite{liu2019naomi} were proposed to apply forward and backward RNN on observed points to infer missing values in each trajectory. However, those methods still rely on learning sequential dependencies of points in trajectories, and the experiments only tested the trajectories of agents in a relatively small scene (e.g., a billiard table or basketball square). The performance of daily human movements' trajectories across city regions remains unknown. This type of trajectory contains more arbitrary movements occurring in an ample space with possibly more sparsity. Our approach can provide insights for capturing the coexisting dependencies between missing positions and observed ones on different human mobility datasets. And the model is established based on the conditional independence assumption, considering both spatial and temporal perspectives, which the previous work did not examine sufficiently.

\section{Problem Definition}
\label{sec:problem_def}

\begingroup
\setlength{\tabcolsep}{6pt}
\begin{table}[tb]
\centering
\caption{Symbols and notations.}
\begin{tabular}{cl} 
\hline\hline
\textbf{Symbols} & \textbf{Description} \\ 
\hline
$p_t$ & The point was recorded at time $t$  \\
$S_u$ & The original mobility sequence of user $u$  \\
$tr_u^w$ & One sub-trajectory of $S_u$ in $w$-$th$ time window  \\
$U$ & The number of users involved \\
$T$ & The maximum number of points considered  \\
$L$ & The width of each time window  \\
$S_u^{'}$ & A set of sub-trajectories of user $u$ from $S_u$  \\
$p_l$ & The point at step $l$ in a trajectory \\
$p_i$ & A missing point in one trajectory \\
$M$ & A masking vector for missing values \\
$I$ & The total number of missing points in trajectory \\
$D$ & The dimension of initial representation \\
$Z$ & An initial representation of input points\\
$z_{obs}^l$ & The initial representation of point $p_l$ \\
$z_{mis}^i$ & The initial representation of missing point $p_i$ \\
$\lambda_1$ & The weight of imputation component\\
$\lambda_2$ & The weight of prediction component\\
$\lambda_3$ & The weight of movement velocity\\
\hline
\end{tabular}
\label{fig:symbols}
\vspace{-5pt}
\end{table}
\endgroup

In this section, we define the problem of trajectory prediction and imputation task as follows: 
\begin{itemize}
    \item \textbf{The Prediction Task:} we denote an original mobility sequence of user $u$ as $S_u = \{p_1, p_2, . . . , p_T\}$, where $p_t \in \mathbb{R}^2$ is the coordinates of point recorded at time frame $t$ and $T$ is the maximum number of observed values in consideration. A trajectory $tr_u^w$ is one sub-sequence of $S_u$ in $w$-$th$ time window, and the width of the window is $L$. Therefore, $tr_u^w = \{p^{w}_{1}, p^{w}_{2}, . . . , p^{w}_{L}\}$, $l$-$th$ is the number of points in the sub-sequence. A set of such trajectories can be extracted from the original sequence with a defined time window for user $u$: $S_u^{'} = \{tr_u^1, tr_u^2, . . .tr_u^W\}$. The whole dataset can be denoted as $S = \{S_1^{'},S_2^{'}, . . . S_U^{'}\}$ and $U$ is the total number of users. The goal of this task is to predict the coordinates of the next movement when giving a trajectory.
    \item \textbf{The Imputation Task:} we assume that $tr_u^w$ denotes one of user $u$'s trajectories. In reality, $tr_u^w$ may contain a portion of missing points for many reasons. Thus, the states of all the points inside the trajectory are represented with one masking vector $M = [m_1, m_2, . . ., m_L]$, where $m_l$ equals to zero if $p^{w}_{l}$ is not observed. Otherwise, $m_l$ is set to one. The purpose is to infer and substitute the missing values with appropriate alternatives in each trajectory.
\end{itemize}
Moreover, Table \ref{fig:symbols} provides more information about the symbols and notations used in this paper.

\section{Proposed Approach}
\label{sec:proposed_work}
\begin{figure}[!t]
\centering
\includegraphics[width=0.8\linewidth]{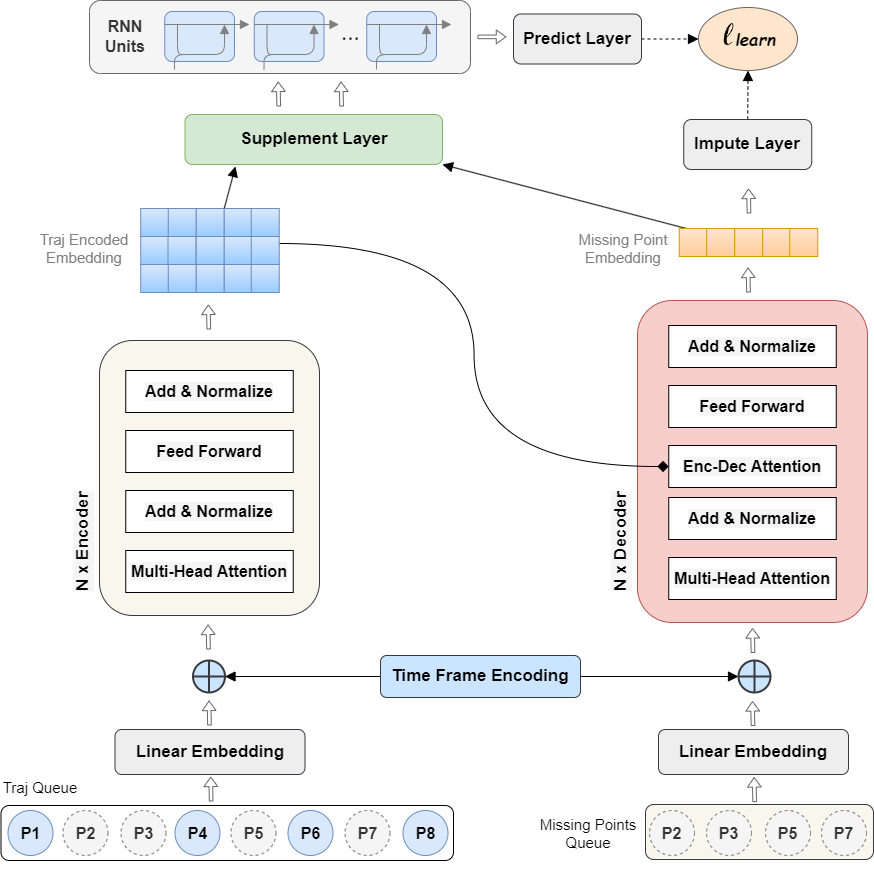}
\caption{The proposed framework solves trajectory imputation and prediction jointly. $P1, P4, P6$, and $P8$ are the observed points in the trajectory, $P2, P3, P5$, and $P7$ are the missing values. Multiple Encoders and Decoders are employed in the learning process to produce high-level embeddings for the observed trajectory and missing values, respectively. First, the Linear embedding layer initializes the representations of the trajectory queue and missing point(s), and Time Frame Encoding is responsible for adding observed time points to relevant representations. Next, the trajectory embeddings from Encoders are fed to Encoder-Decoder Attention to enhance the embedding of missing points. Finally, new embeddings of the trajectory are reconstructed by the Supplement Layer and transferred to an RNN-based unit for prediction purposes. In addition, the Imputation Cycle enables the model to conduct gradual imputation on multiple levels.}
\label{fig:main-frame}
\end{figure}

Fig.~\ref{fig:main-frame} illustrates the architecture of our proposed method for both human trajectory imputation and prediction. For imputation purposes, encoders and decoders based on self-attention are applied to learn coexisting patterns between observations and missing points. Then, a recurrent component plays a role in extracting sequential dependencies on newly learned embeddings from the Supplement Layer for forwarding prediction. Meanwhile, a mechanism called the Imputation Cycle is introduced to achieve progressive learning of both imputation and prediction at multiple levels. The model interactively considers learning two main components and improves overall performance. Algorithm \ref{alg:Main_alg} shows an overview of training conducted using the proposed approach, and the details of the main components are described in the following subsections.

\subsection{Imputation Component}
In the beginning, the representations of each trajectory with missing points are generated via Linear layers and Frame-positional Encoding. Then, the imputation component takes the initial representations and captures the spatial-temporal dependencies among points through self-attention in either Encoders or Decoders. This process will produce embeddings for observations and missing values, respectively. Here, one agile design allows the model to decode $n$ number of missing points at each imputation time, and the process is repeated by the Imputation Cycle. Thus, the final impute layer is responsible for inferring $n$ number of missing points each time.

\subsubsection{Feature Initialisation} 
In the first stage, the framework projects representations of an incomplete trajectory into a higher $D$-dimensional space via the Linear Embedding Layer. Specifically, the initial representation of one observed point $p_l$ is $z_{obs}^l = p_l^{\top} W_{obs}$, where $\top$ denotes transpose matrix, $W_{obs}$ is the weight matrix, and $p_l$ is a zero vector if it is missing. We obtain a queue for the incomplete trajectory $E_{obs}=\{z_{obs}^1,z_{obs}^2,...,z_{obs}^L\}$. We extract another queue of missing points from $E_{obs}$ as $E_{mis}=\{z_{mis}^1,z_{mis}^2,...,z_{mis}^I\}$. Similarly, the initial representation of a missing point is $z_{mis}^i = p_i^{\top} W_{mis}$, where $p_i$ is also zero vector $(0 \leq i \leq I)$ and $I$ is the total number of missing points in a trajectory.

Adding time information to the initial representations of trajectories is essential because the imputation unit relies on the attention mechanism for learning without any sequential knowledge. We insert time representation to the points as distinctive identifications in trajectory based on the Positional Encoding method \cite{vaswani2017attention}. The relevant equations are given as follows:
\begin{equation}
F(t,d)=\left\{\begin{matrix}
\sin(\frac{t}{10000^{d/D}}), & \textit{when d is even,}\\ 
\cos(\frac{t}{10000^{d/D}}), & \textit{when d is odd,}
\end{matrix}\right.
\end{equation}
where $F(t,d)$ outputs the representation vector for time frame $t$ recorded for a point in the trajectory, and $d$ is the $d$-th dimension in the vector, which is also $D$-dimensional. Then, the representation of the time frame is added to (e.g., element-wise addition) the initial embedding of the corresponding point in either $E_{obs}$ or $E_{mis}$. Noticeably, the $F$ is applied to the missing points in both queues $E_{obs}$ and $E_{mis}$, such that the model can have consistent temporal information of missing values from both queues during the imputation.

The original Positional Encoding encodes the positions of words in a sequence. Here, we encode the time frames of points in a trajectory within the observed period. One time frame $t$ is a unique numeric index like $0, 1, 2...$, which stands for the order in which people's movements occurred in the observation period (day, week, or month). Some existing literature \cite{feng2018deepmove} has mentioned that the patterns of human mobility tend to occur periodically, and integrating with this information can help to learn relevant tasks. Therefore, we add time information using the time positional encoder to embed the time frame of points during a considered period. Although the missing points are initially represented with zero vectors, the Positional Encoding attaches time features on each missing value which will be further updated by the attention-based Encoders and Decoders in the next training stage.

\subsubsection{Encoder and Decoder} 
As Fig. \ref{fig:main-frame} shows, multiple Encoders and Decoders are employed to produce embeddings by weighting the correlation between observed and missing values, respectively. They share a similar structure: Self-attention, Normalisation, and Feed-forward Layers. The self-attention block of the first encoder and decoder acquires $E_{obs}$ and $E_{mis}$ as inputs, respectively. And they are responsible for obtaining the internal dependencies among the points on each side. Moreover, each decoder contains an Encode-decode Attention to learn further external dependencies between an incomplete trajectory and its missing values. More specifically, a $d_k$-dimensional $query$ vector, $d_k$-dimensional $key$ vector, and $d_v$-dimensional $value$ vector are built for each point by multiplying its embedding with three different weight matrices, respectively. In practice, we deal with a set of points together by wrapping the $query$ vectors into matrix $Q$, $keys$ into matrix $K$, and $values$ into matrix $V$. The final output of the self-attention layer is calculated using the following formula:
\begin{equation}
Attention(Q,K, V)=softmax(\frac{QK^\top}{\sqrt{d_k}})V,
\label{equ_self_att}
\end{equation}
where $d_k$ is the dimension of $K$. Multiple encoders or decoders work sequentially in the imputation component: the output of one block will be taken as input by the next block. 

According to \cite{vaswani2017attention}, multi-head attention that comprises several self-attentions can be applied to jointly synthesize the initial representation $Z$ of points in each input queue from different representation sub-spaces, which helps enhance embedding learning. The functions of the operation are given as follows:
\begin{equation}
\begin{aligned}
MultiHead(Z) & = Concat(head_1, head_2,..., head_h)W^O, \\  
& W^O \in \mathbb{R}^{hd_v \times D},
\label{equ_mul_head_att}
\end{aligned}
\end{equation}
\begin{equation}
\begin{aligned}
head_i = Attention(ZW^Q_i, ZW^K_i, ZW^V_i), \\
W_i^Q \in \mathbb{R}^{D \times d_k}, W_i^K \in \mathbb{R}^{D \times d_k}, W_i^V \in \mathbb{R}^{D \times d_v}.
\label{equ_head_att}
\end{aligned}
\end{equation}
Here, $h$ is the total number of heads in consideration, and each $head_i$ is an individual of self-attention. $Q_i=ZW_i^Q, K_i=ZW_i^K$ and $V_i=ZW_i^V$. One of the main merits of multi-head attention is that attending computations can be executed in parallel to boost the overall runtime. We apply two heads of self-attention in both encoders or decoders in the experiments.

\begin{figure}[!t]
\centering
\includegraphics[width=0.56\linewidth]{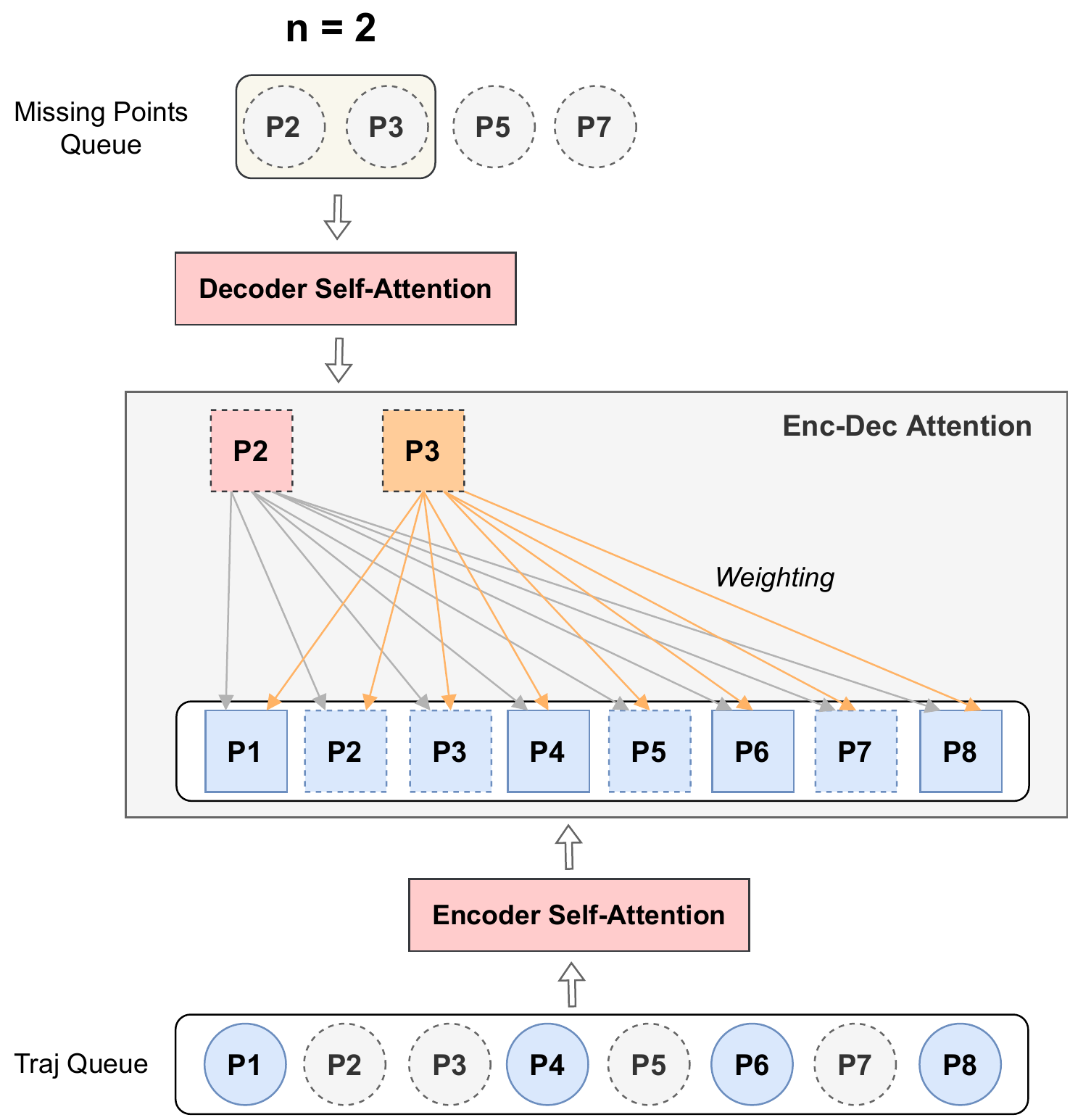}
\caption{The diagram shows that the Encoder-Decoder Attention captures the dependent patterns between the missing points and an observed trajectory.}
\label{fig:enc-dec-atten}
\end{figure}

The proposed framework learns the coexistence patterns of the observed and missing points by capturing their dependencies between the point embeddings from the encoders and decoders. Fig. \ref{fig:enc-dec-atten} illustrates the component of an Encoder-Decoder Attention that extracts the dependent relations between the observed trajectory and missing points. For the imputation of one incomplete trajectory, the queues are created for missing points and the relevant trajectory, respectively. Furthermore, the model imputes $n$ missing points each time from the queue of missing points. In the example of Fig. \ref{fig:enc-dec-atten}, when $n=2$, missing points $P2$ and $P3$ are first ejected into the Decoder Self-Attention for producing embeddings of $P2$ and $P3$. Simultaneously, the queue of the incomplete trajectory is injected into the Encoder Self-Attention for trajectory embedding production. Then, the Encoder-Decoder Attention is responsible for learning the dependencies by weighting the relationship of missing points ($P2, P3$) and the observed trajectory. Finally, the locations of the missing points can be inferred based on their learned embeddings.

\subsubsection{Imputation Recurrence}
\begin{figure}[!t]
\centering
\includegraphics[width=0.8\linewidth]{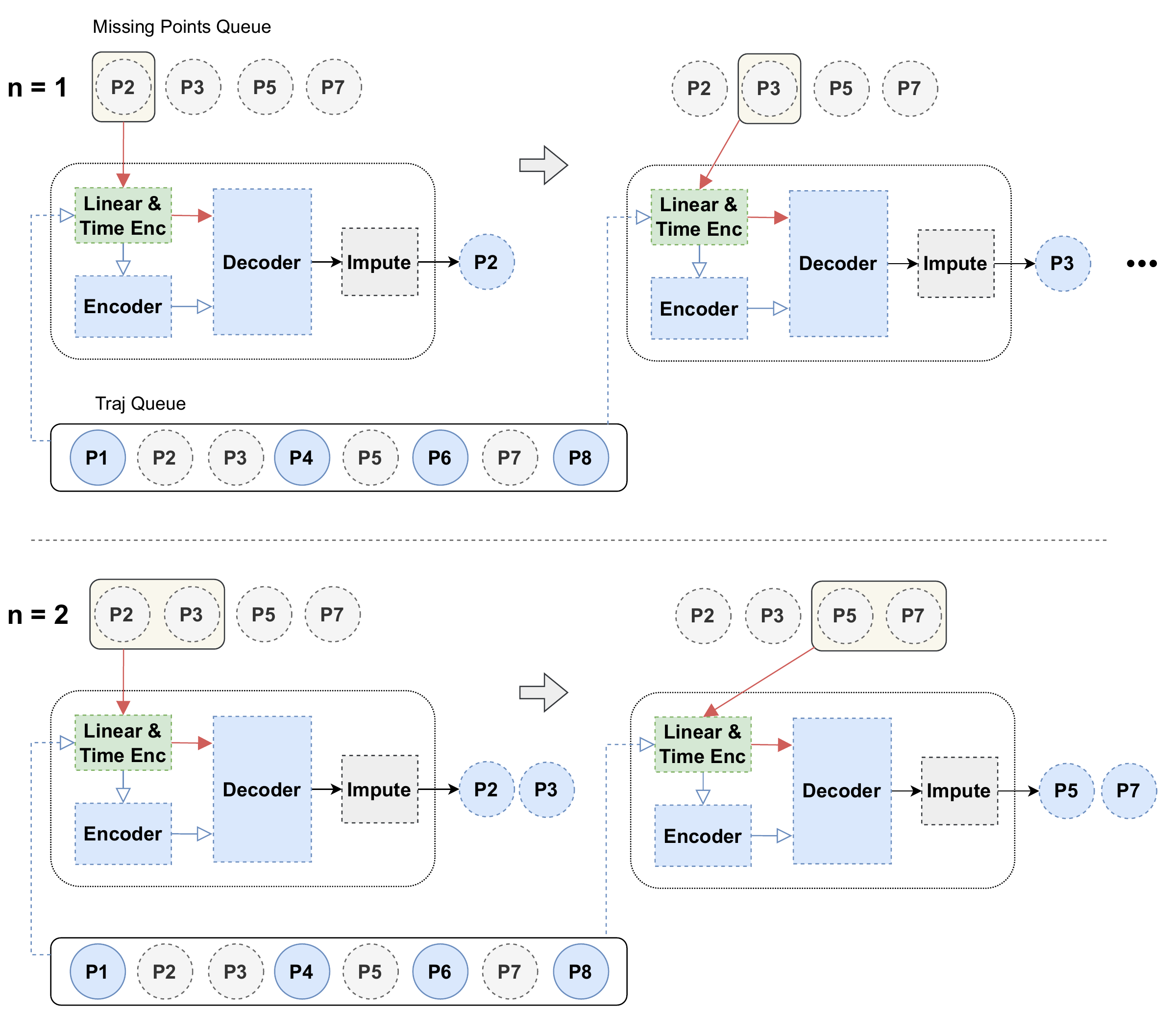}
\caption{The Imputation Cycle enables the model to conduct gradual imputation on multiple levels. In each imputation cycle, $n$ missing points will be taken into the model for learning with the relevant trajectory information. The diagram shows the imputation process when $n$ equals $1$ or $2$.}
\label{fig:point-impute-cycle}
\end{figure}

The model is capable of carrying out gradual imputation at multiple levels, which gives the model more flexibility by inferring $n$ $(1<n<I)$ missing nodes in each imputation cycle. The encoders yield the embeddings of an incomplete trajectory and then transfer them to decoders for weighting with missing values in imputation. Before that, the decoder takes the $n$ number of missing values as inputs in each cycle. Further, $n$ alternatives are yielded from the Imputing Linear Layer, and new learning circulation is started based on the previous state. The model can train and learn on each trajectory progressively. As Fig. \ref{fig:point-impute-cycle} shows, the model takes $n$ points from the missing points queue in each imputation cycle. In a decoder, the embeddings of $n$ missing points are weighted based on observed trajectory embedding produced by an encoder. Then, the impute layer outputs the locations of $n$ missing points. The imputation cycle will be repeated until all the missing points have been learned.

\subsection{Prediction Component}
The details of the prediction component are provided in this section. At the upper part of the framework in Fig. \ref{fig:main-frame}, it is the Supplement Layer that fetches the embeddings of observed portion $Y_{obs}$ and missing values $Y_{mis}$ from the final layer of encoder and decoder, respectively. The Supplement layer is responsible for reconstructing a new trajectory sequence by replenishing the trajectory representation with the embeddings of the missing points after the encoding stage. Subsequently, an RNN-based unit is built upon this module to capture sequential dependency. The formula of a supplement operation is given below:
\begin{equation}
\begin{aligned}
Y_{seq}(p_{l})=\left\{\begin{matrix}
Y_{mis}(m_{i}), & \textit{if } p_{l} = m_{i}\\ 
Y_{obs}(o_{i}), & \textit{if } p_{l} = o_{i}
\end{matrix}\right.
\label{equ_supp_emb}
\end{aligned}
\end{equation}
where $Y_{seq}$ is the embedding matrix of the whole trajectory, $p_{l}$ is a point at step $l$ in the trajectory, $m_{i}$ is a point in the missing queue, and $o_{i}$ denotes a point in the observation queue. We interpolate the embedding of a point from matrix $Y_{mis}$ if it is missing; otherwise, the corresponding embedding is extracted from matrix $Y_{obs}$. In addition to using a 'Replace operation', we can alternatively consider an 'Add operation' in the Supplement layer to add the embedding of missing values to the corresponding positions in the trajectory representation.

In the prediction stage, GRU \cite{chung2014empirical} is selected as the recurrent unit because of its efficient computation without performance deterioration. This layer is implanted by receiving the embeddings of the newly reconstructed sequence from the supplement layer and produces relevant hidden states. The updated formulations of GRU are as follows:
\begin{equation}
f_t = \sigma(W_{fx}x_t + W_{fh}h_{t-1} + b_f),
\label{equ_gru_1}
\end{equation}
\begin{equation}
r_t = \sigma(W_{rx}x_t + W_{rh}h_{t-1} + b_r),
\label{equ_gru_2}
\end{equation}
\begin{equation}
c_t = tanh(W_{cx}x_t + r_t \circ W_{ch}h_{t-1} + b_c),
\label{equ_gru_3}
\end{equation}
\begin{equation}
h_t = (1-f_t) \circ c_t + f_t \circ h_{t-1},
\label{equ_gru_4}
\end{equation}
where $x_t$ is the embedding of the point in time $t$, $h_{t-1}$ is the output of the last unit, $W$ is the weight matrix, $b$ is the bias vector, $\circ$ means element-wise multiplication, $f_t$ is the update gate, $r_t$ is the reset gate, $c_t$ is the candidate, and $h_t$ is the output. Noticeably, the model flexibly subjoins prediction training in each imputation cycle. In other words, it can recurrently learn and make forward predictions based on the latest status of the sequence restored by previous imputation recurrences. In the testing step, we need to detach the forward prediction from the imputing cycles and solely execute it when the imputation of the whole trajectory is completed.

\begin{algorithm}[!tb]
	\caption{Overview of INGRAIN Training}
	\begin{algorithmic}[1] 
	\vspace{0.5mm}
		\Require 
		\Statex A set of user trajectory $S_u^{'} = \{tr_u^1, tr_u^2, . . .tr_u^W\}$; 
		\Statex One masking vector $M = [m_1, m_2, . . ., m_L]$; 
		\Statex Epochs of training $epo$;
		\Statex Number of points for each imputation cycle, $n < I$.
		\vspace{0.5mm}
		\While{training epoch $< epo$}
    	    \ForAll{trajectory $tr_u^w \in S_u^{'}$}
                \State Apply mask $M$ on $tr_u^w$;
                \State Initialize representations of observed trajectory, $E_{obs}$;
                \State Encode time information with $E_{obs}$;
    		    \While{has missing values}
    		        \State Initialize representations of $n$ missing points, $E_{mis}$;
    		        \State Apply time encoding on $E_{mis}$;
    		        \State Compute attention for observations, $E^{'}_{obs}$;
    		        \State Compute attention of $n$ missing values in observations, $E{'}_{mis}$;
    		        \State Reconstruct trajectory based on $E^{'}_{obs}$ and $E{'}_{mis}$;
    		        \State Conduct imputation and prediction;
    		        \State Optimisation with fused objective function $\mathcal{L}_{learn}$.
                \EndWhile 
    		\EndFor
		\EndWhile
	\end{algorithmic} 
	\label{alg:Main_alg}
\end{algorithm} 

\subsection{Collaborative Learning Objective}
\label{sec:collaborative_learning_objective}
As we mentioned, there are two main components established in the framework: imputation and prediction units. During the training stage of imputation, the purpose is to minimize the mean squared error between imputing values and ground truth with the following objective function:
\begin{equation}
\mathcal{L}_{imp} = \mathbb{E}_{X \sim C, \widehat{X} \sim G_\theta(X,M)}
\begin{bmatrix}
\sum_{l=1}^{L} 
\begin{Vmatrix}
\hat{x_l} - x_l
\end{Vmatrix}_2
\end{bmatrix},
\label{equ_obj_imp}
\end{equation}
where $C=\{X^*\}$ is the set of original sequences, $M$ is a masking vector for missing values, $G_\theta(X,M)$ denotes the function to infer missing values for imputation with parameter $\theta$, and $\hat{x}$ is one of the imputed values $\widehat{X}$. Subsequently, we generate embedding $Y$ of the sequence from the previous component in the training process of prediction. $J_\omega(Y)$ is the predictor of the next movement with parameter $\omega$, and the objective function of the prediction task is constructed as follows:
\begin{equation}
\mathcal{L}_{pre} = \mathbb{E}_{X \sim C, Y \sim G_\theta(X,M), \hat{y} \sim J_\omega(Y)}
\begin{bmatrix}
\begin{Vmatrix}
\hat{y} - y
\end{Vmatrix}_2
\end{bmatrix}.
\label{equ_obj_pre}
\end{equation}
 Furthermore, we introduce a third loss function $\mathcal{L}_{vel}$ to constraint the movement velocity between imputed and observed points and comply with the speed observed in trajectories:
\begin{equation}
\mathcal{L}_{vel} = \mathbb{E}_{X \sim C, \widehat{X} \sim G_\theta(X,M), v \sim H(X - \widehat{X}), \hat{v} \sim H(\widehat{X})}
\begin{bmatrix}
\begin{Vmatrix}
\hat{v} - v
\end{Vmatrix}_2
\end{bmatrix},
\label{equ_obj_vel}
\end{equation}
where $H$ is the function to compute movement speed between each pair of points, $v$ is the observed speed in a trajectory, and $\hat{v}$ is the speed computed between imputed and observed points.

To train the entire model, we fuse the optimization of the above objective functions in each execution as shown below:
\begin{equation}
\mathcal{L}_{learn} = \lambda_1 \mathcal{L}_{imp} + \lambda_2 \mathcal{L}_{pre} + \lambda_3 \mathcal{L}_{vel}.
\label{equ_obj_learn}
\end{equation}
Here, $\lambda_1$,  $\lambda_2$, and $\lambda_3$ are the hyperparameters representing the weights of different loss functions correspondingly in training. In this way, the model can benefit from optimizing different modules, and this collaborative learning could eventually give the model a latent boost of convergence.

\subsection{Computational Complexity}
\label{sec:complex_ana}
In this part, we take into account the main phases of the proposed framework for calculating computational complexity. The meanings of symbols used here are independent of the notations in the previous sections. The essential phases include initial representation generation, encoder and decoder attention, and RNN-based prediction module.

The stage of initial representation generation is directly built on Multi-Layer Perceptron (MLP) \cite{khanna1990foundations}, which approximately has time complexity $O(l*n*d)$ for one layer of implementation. Here, $l$ is the trajectory length, $n$ is the input dimension (it is regularly a small value), and $d$ is the embedding dimension. In the imputation stage, the encoder and decoder mainly rely on the self-attention mechanism, which has $O(l^{2}*d)$ \cite{vaswani2017attention}. On the other hand, the RNN-based module typically contributes to time complexity $O(l*d^{2})$. In most cases, the trajectory length $l$ is smaller than the embedding dimension $d$. However, self-attention is unnecessary to conduct sequential operations on all the points. It is able to consider neighbor locations of size $r$ in the input trajectory if it is particularly long \cite{vaswani2017attention}. So, the complexity could reduce to $O(l*r*d)$, which will be considered in future work. Overall, the 
$O(max\{l*n*d, l^{2}*d, l*d^{2}\})$ is the total complexity of the framework. In the experiments, $n$ is two for the input size, and $l$ and $d$ are configured correspondingly to different values.

\section{Experimental Results}
\label{sec:experiment}
This section compares the performance of trajectory imputation and prediction for the proposed model and the baselines on different human mobility datasets. Then, the primary hyperparameters of the proposed model are assessed intensively. A ablation study is provided at the end of the section.

\subsection{Experimental Setup}
\subsubsection{Datasets}
We use real-world human mobility datasets from three different cities worldwide for the experiments. The datasets record a broad range of users’ movements among city areas:
\begin{itemize}
\item \textbf{Geolife Data} \cite{zheng2009mining} contains outdoor GPS trajectories of 182 users from April 2007 to August 2012 in Beijing, China. The sampling rates vary in trajectories: approximately $91\%$ are logged every 1 to 5 seconds or every 5 to 10 meters per point. Each record contains \textit{timestamp, user ID, latitude and longitude}. We selected 30 users with the most GPS records in January and February 2009 for evaluation.
\item \textbf{Cuebiq-US Data} \cite{cuebiq2021} contains more diverse human movements on a daily basis. The data collection period ranged from January 2018 to June 2018, and the location was New York, USA. Sampling frequencies of approximately $91\%$ of data range from 1 to 600 seconds per record, and each record has \textit{timestamp, device ID, latitude, and longitude}. Cuebiq's anonymized and privacy-enhanced data is collected from users who opted for anonymous data sharing for research purposes through a GDPR-compliant framework. The trajectories of the 30 most active users in May 2018 are extracted for experiments.
\item \textbf{Cuebiq-AU Data} \cite{cuebiq2021} has the same data format as Cuebiq-US. It was collected over two years, from December 2017 to November 2019, in cities in Australia. The sampling rate of the collected data is similar to that of Cuebiq-US for most records. Likewise, The trajectories of 30 users who were most active in October 2019 are used for testing.
\end{itemize}

\begingroup
\setlength{\tabcolsep}{6pt}
\begin{table}[t]
\centering
\caption{The total number of trajectories used in each dataset and $L$ denotes the length of each trajectory.}
\label{tab:dataset_traj_num}
\begin{tabular}{ccc|ccc|ccc} 
\hline\hline
\multicolumn{3}{c|}{\textbf{Geolife}}          & \multicolumn{3}{c|}{\textbf{Cuebiq-AU}}        & \multicolumn{3}{c}{\textbf{Cuebiq-US}}          \\ 
\hline
\textit{L} = 20 & \textit{L} = 50 & \textit{L} = 100 & \textit{L} = 20 & \textit{L} = 50 & \textit{L} = 100 & \textit{L} = 20 & \textit{L} = 50 & \textit{L} = 100  \\ 
\hline
20,435        & 20,435        & 20,301         & 30,030        & 30,030        & 30,030         & 30,030        & 30,030        & 30,030          \\
\hline
\end{tabular}
\end{table}
\endgroup

Based on the original movement sequences of the users in each dataset, we produce a set of sub-trajectories of the users with a defined length (see Section \ref{sec:problem_def}). Table \ref{tab:dataset_traj_num} gives the total number of trajectories used in each dataset by length. Then, the extracted trajectories are randomly separated into a training part ($80\%$) and a testing part ($20\%$). Additionally, a varied size of masking vector is randomly created to imitate different degrees of missing values in sub-trajectories for imputation. The default probability of generating missing values follows a discrete uniform distribution. Finally, the model requires predicting the location of the next movement when an observed trajectory with missing points is given.

\subsubsection{Metrics} In this paper, the losses of $\mathcal{L}_{imp}$ and $\mathcal{L}_{pre}$ are basically average euclidean distances between 2D points. In imputation, we infer the coordinates of missing points in trajectories and calculate the average $L2$ loss between imputed values and ground truth among all users. Moreover, we assess the proposed approach for the prediction task, which is to forecast the coordinates of the next location for users if a historical trajectory with missing points is given. The average $L2$ loss between the predicted values and ground truth is used for the evaluation. In our model, the predicted values will be evaluated merely after a trajectory's imputation is completed during the testing phase.

\subsubsection{Baselines} For trajectory imputation, two state-of-the-art methods for comparison are \textbf{NAOMI} \cite{liu2019naomi} and \textbf{SingleRes} \cite{liu2019naomi}. NAOMI is one of the latest non-autoregressive approaches for sequence imputation. In contrast, SingleRes is the autoregressive counterpart, and it can be reduced to BRITS \cite{cao2018brits} if the adversarial training is discarded. \textbf{GRUI} \cite{luo2018multivariate} is also an autoregressive model with GAN for time series imputation, which is used to handle the completion of a trajectory sequence. Moreover, \textbf{GAIN} \cite{yoon2018gain} is another recent method to impute missing data using GANs. We also test the imputation task with a classical approach named \textbf{KNN + Linear} \cite{hastie2009elements}, which searches K nearest neighbors from samples and then applies Linear regression to impute missing points based on those neighbors. Here, we also implemented this approach by inferring a defined number of missing points in a trajectory in different degrees. Simultaneously, several RNN variations are used for comparison with the proposed model for prediction purpose, such as stacked LSTM (\textbf{S-LSTM}) \cite{sundermeyer2012lstm}, bidirectional LSTM (\textbf{B-LSTM}) \cite{graves2005bidirectional} and stacked GRU (\textbf{S-GRU}) \cite{chung2014empirical}. Another sequence forecasting method is \textbf{RNNSearch} \cite{bahdanau2014neural}, which implements the attention mechanism based on RNN to selectively retrieve information from the encoder to the decoder for prediction.

\subsubsection{Implementation}
The proposed method (INGRAIN) is implemented with Pytorch, and the Adam algorithm is used as the optimizer with a learning rate of $0.001$ and batch size of $70$. In the imputation component, we use two layers of either encoders or decoders. The number of heads (self-attention) in each layer is two, and the dimension of learning embedding is $256$. In the prediction part, 1-layer GRU is adopted with a hidden size of $256$. For training in different settings, the number of epochs is 60, and we compute the mean of the best test results for each task in five runs.

For other imputation methods, NAOMI and SingleRes \cite{liu2019naomi} use the same values of some basic parameters as our model: learning rate 0.001, batch size 70, and training epochs 60. The other parameters are default in their implementation. GRUI \cite{luo2018multivariate}, and GAIN \cite{yoon2018gain} are faster for training but harder to achieve convergence. Furthermore, we tried a different number of iterations for training to obtain optimal results on different datasets, ranging from 50 to 1000. The baselines are all RNN based for the prediction methods, and we can directly compare them with our prediction component by applying similar parameters for training, such as learning rate, batch size, training epochs, or size of hidden features.

\subsection{Performance Analysis}
\label{perform_analysis}
\begingroup
\setlength{\tabcolsep}{6pt}
\begin{table*}[!tb]
\centering
\caption{The results show the L2 loss of imputation and prediction on three different datasets. The percentage of missing points in each trajectory is 0.8 for all the tests, and $L$ denotes the length of trajectories in different trials.}
\label{tab:imp_pre_loss}
\begin{tabular}{c|ccc|ccc|ccc}
\hline\hline
\multicolumn{10}{c}{\textbf{L2 Loss for Imputation}}                                                                                                                                 \\ 
\hline
\multirow{2}{*}{\begin{tabular}[c]{@{}c@{}}\textbf{ Methods}\\\end{tabular}} & \multicolumn{3}{c|}{\textbf{Geolife }}              & \multicolumn{3}{c|}{\textbf{Cuebiq - AU}}           & \multicolumn{3}{c}{\textbf{\textbf{Cuebiq - US}}}    \\ 
\cline{2-10}
                                                                             & \textit{L} = 20   & \textit{L} = 50   & \textit{L} = 100  & \textit{L} = 20   & \textit{L} = 50   & \textit{L} = 100  & \textit{L} = 20   & \textit{L} = 50   & \textit{L} = 100   \\ 
\hline
KNN + Linear \cite{hastie2009elements}                                                                 & 4.7095          & 4.7014          & 8.7260          & 1.6466          & 1.9836          & 1.9567          & 0.0238          & 0.0254          & 0.0323           \\
GAIN \cite{yoon2018gain}                                                                         & 2.1597          & 2.1903          & 2.9478          & 0.8619          & 0.8842          & 1.0788          & 0.0125          & 0.0126          & 0.0137           \\
GRUI \cite{luo2018multivariate}                                                                         & 0.3884          & 0.2584          & 0.2443          & 0.3150          & 0.2062          & 0.2088         & 0.8407          & 0.2871          & 0.2189           \\
NAOMI \cite{liu2019naomi}                                                                        & 0.0498          & 0.1613          & 0.0598          & 0.2007          & 0.0186          & 0.0524          & 0.0122          & 0.0129          & 0.0127           \\
SingleRes \cite{liu2019naomi}                                                                    & 0.4365          & 0.0405          & 0.0171          & 0.0716          & 0.0190          & 0.0622          & ~0.0121         & 0.0128          & 0.0126           \\
INGRAIN (ours)                                                           & \textbf{0.0270} & \textbf{0.0075} & \textbf{0.0062} & \textbf{0.0122} & \textbf{0.0116} & \textbf{0.0117} & \textbf{0.0055} & \textbf{0.0050} & \textbf{0.0046}  \\ 
\hline
\multicolumn{10}{c}{\textbf{L2 Loss~for Prediction}}                                                                                                                                                        \\ 
\hline
B-LSTM \cite{graves2005bidirectional}                                                                      & 2.1856          & 2.0427          & 2.4948          & 0.8294          & 0.9935          & 1.0853          & 0.0120          & 0.0127          & 0.0126           \\
RNNSearch \cite{bahdanau2014neural}                                                                    & 2.2282          & 2.0754          & 2.5726          & 0.8809          & 1.0151          & 1.0941          & 0.0962          & 0.0187          & 0.0176           \\
S-GRU \cite{chung2014empirical}                                                                        & 2.2309          & 2.0437          & 2.5305          & 0.8276          & \textbf{0.9919} & 1.0895          & 0.0121          & 0.0129          & 0.0126           \\
S-LSTM \cite{sundermeyer2012lstm}                                                                       & 2.2081          & 2.0656          & 2.5289          & \textbf{0.8257} & 0.9947          & 1.0846          & 0.0120          & 0.0128          & 0.0125           \\
INGRAIN (ours)                                                  & \textbf{0.0525} & \textbf{0.0464} & \textbf{0.0751} & 1.0496          & 1.0139         & \textbf{0.9194} & \textbf{0.0073} & \textbf{0.0070} & \textbf{0.0071}  \\
\hline
\end{tabular}
\end{table*}
\endgroup

\begin{figure*}[tb]
    \centering
    \subfloat[Geolife - Imputation \label{subfig:geo_imp_miss_rate_group}]{%
    \includegraphics[width=0.33\textwidth]{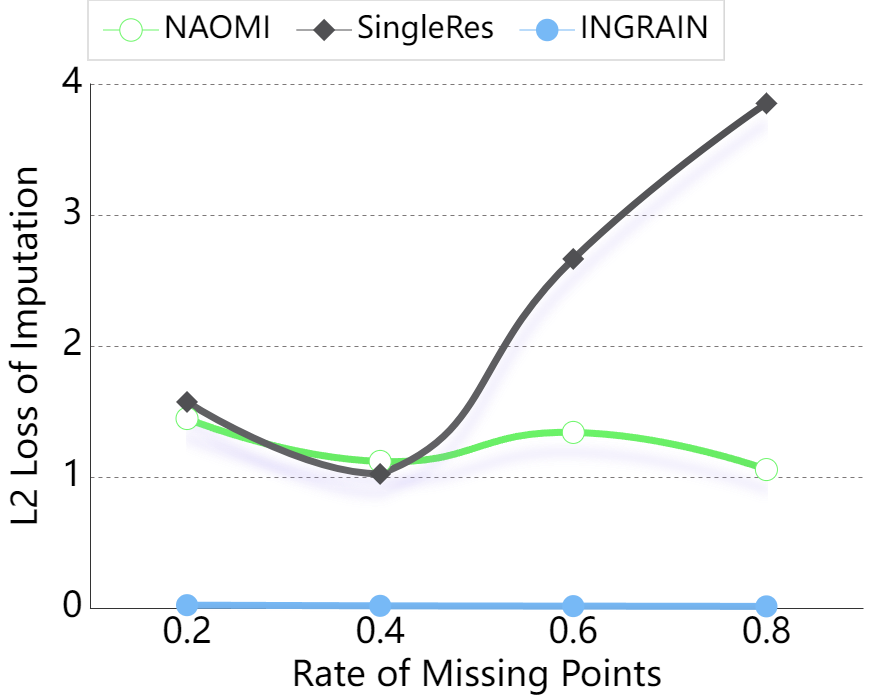}
    }
    \subfloat[Cuebiq-AU - Imputation \label{subfig:cuebiq_au_imp_miss_rate_group}]{%
    \includegraphics[width=0.33\textwidth]{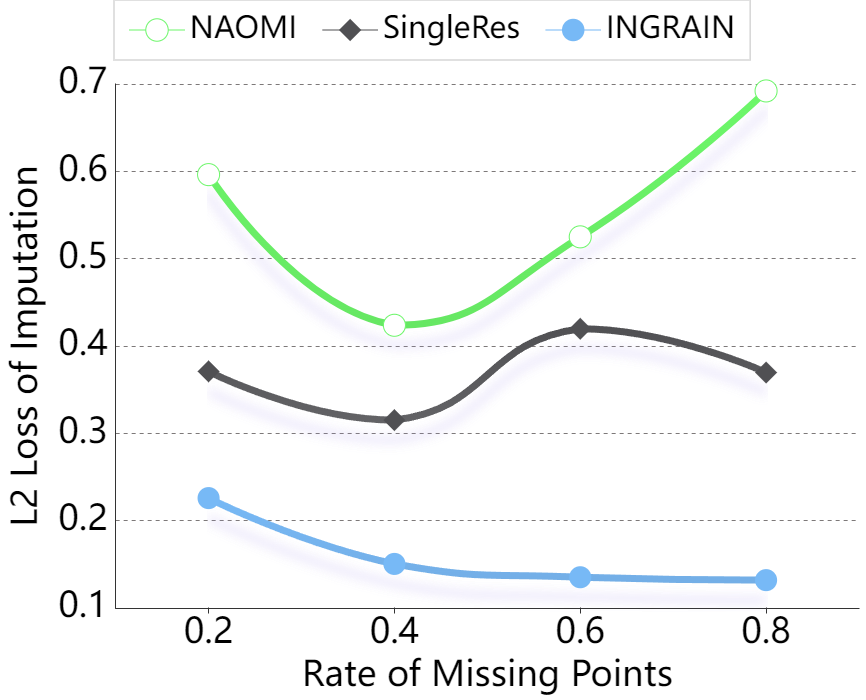}
    }
    \subfloat[Cuebiq-US - Imputation \label{subfig:cuebiq_us_imp_miss_rate_group}]{%
    \includegraphics[width=0.33\textwidth]{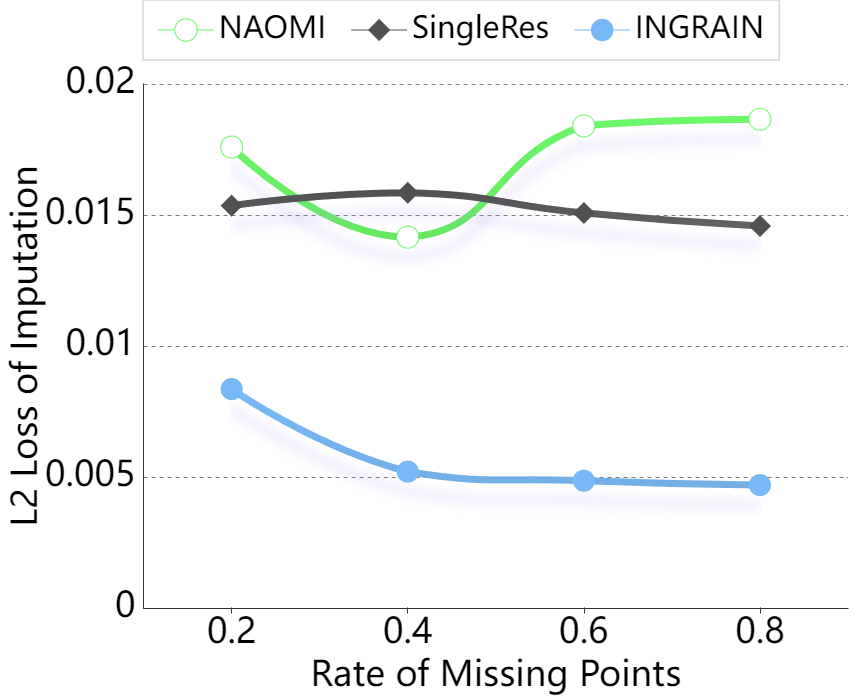}
    } \\
    \subfloat[Geolife - Prediction  \label{subfig:geo_pred_misss_rate_group}]{%
    \includegraphics[width=0.33\textwidth]{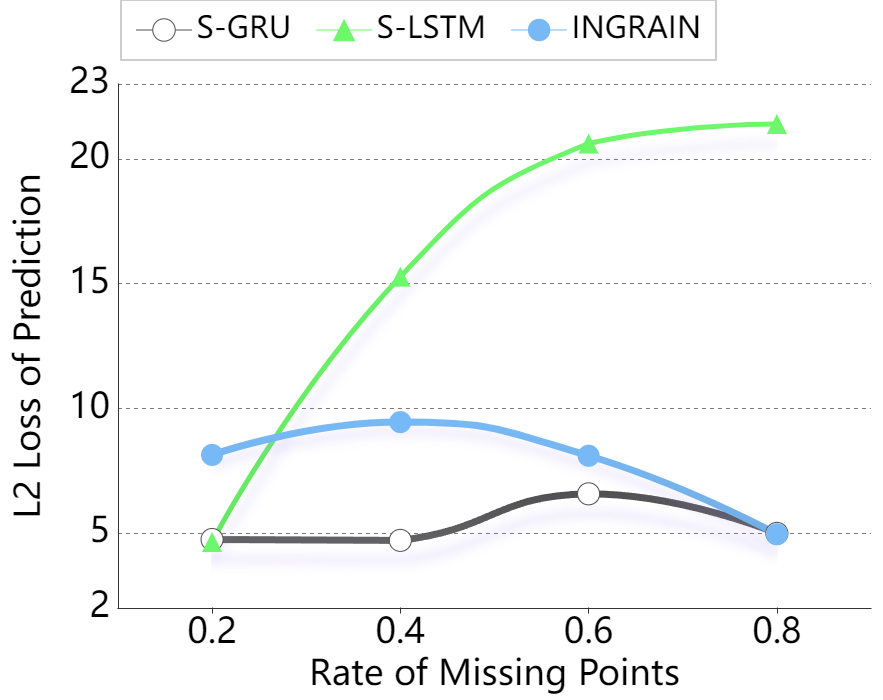}
    }
    \subfloat[Cuebiq-AU - Prediction \label{subfig:cuebiq_au_pred_misss_rate_group}]{%
    \includegraphics[width=0.33\textwidth]{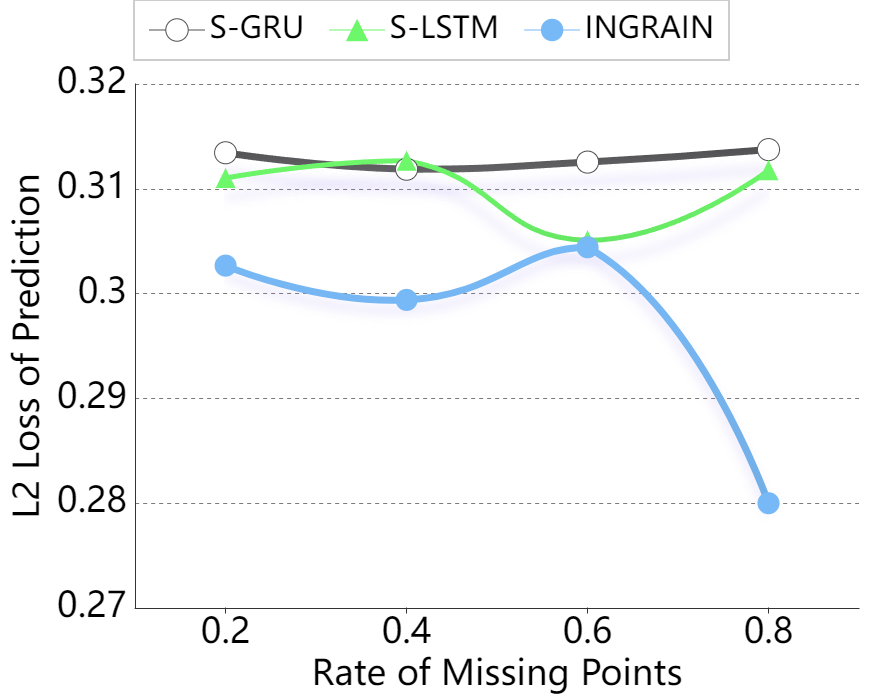}
    }
    \subfloat[Cuebiq-US - Prediction \label{subfig:cuebiq_us_pred_misss_rate_group}]{%
    \includegraphics[width=0.33\textwidth]{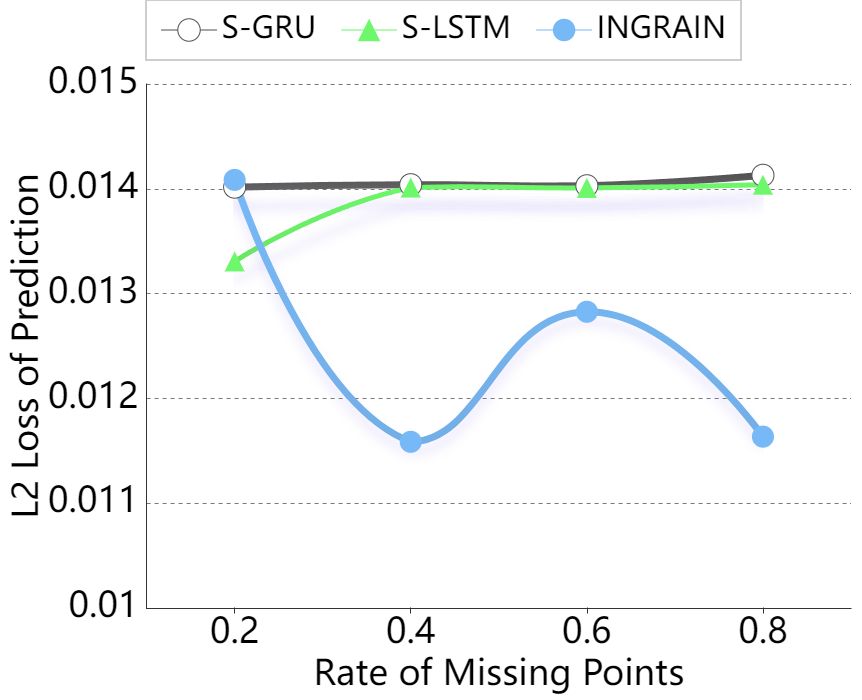}
    }
    \caption{The experiment was run with three groups of 10 users randomly picked from the 30 most active ones in each dataset. The length of a trajectory is 20 in the tests. (a), (b) and (c) demonstrate the imputation loss of the proposed model and baselines on different datasets, with varying percentages of missing points in trajectories. (d), (e) and (f) show the results for prediction loss on three datasets, respectively.}
    \label{fig:imp_loss_group_mis_percent}
\end{figure*}

The evaluations are conducted on Geolife, Cuebiq-AU, and Cuebiq-US, and the sampling rates of points collection vary drastically. This section selected the top 20 users of each dataset who were most active within the observed period for the learning task evaluation, sensitivity analysis, and ablation study. In addition, to further assess the model's effectiveness, we generate three different groups of users, and each group contains ten persons randomly picked from the 30 users in each dataset. The data of groups were tested directly on two learning tasks, and the results are shown in Fig. \ref{fig:imp_loss_group_mis_percent}.

\begin{figure*}[tb]
    \centering
    \subfloat[Geolife - Imputation \label{subfig:geo_imp_miss_rate}]{%
    \includegraphics[width=0.33\textwidth]{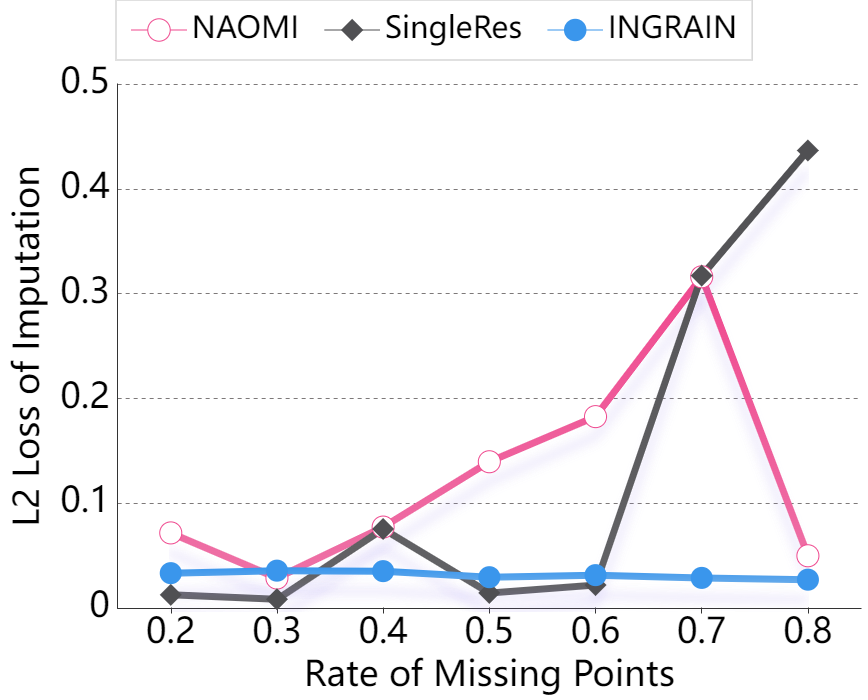}
    }
    \subfloat[Cuebiq-AU - Imputation \label{subfig:cuebiq_au_imp_miss_rate}]{%
    \includegraphics[width=0.33\textwidth]{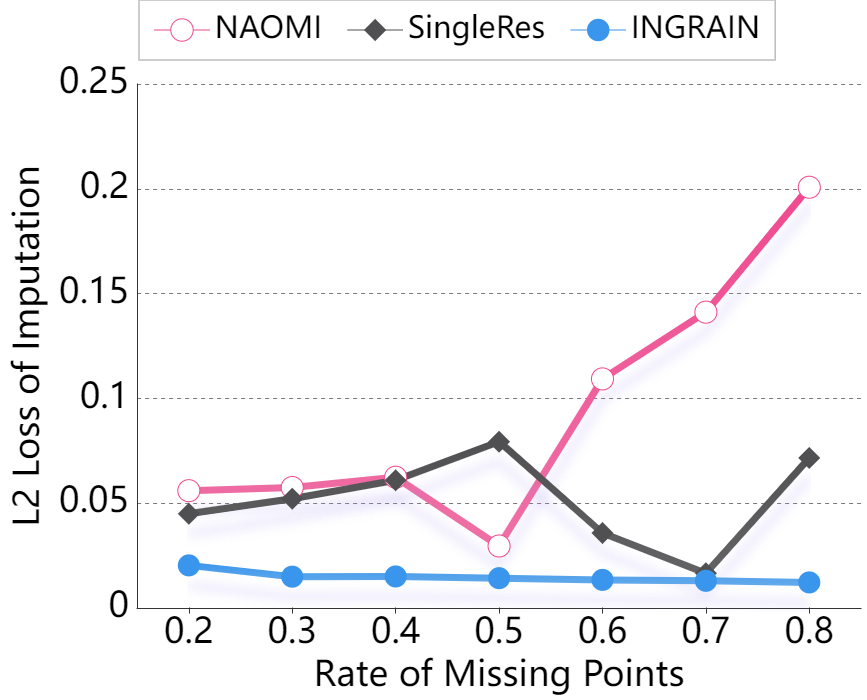}
    }
    \subfloat[Cuebiq-US - Imputation \label{subfig:cuebiq_us_imp_miss_rate}]{%
    \includegraphics[width=0.33\textwidth]{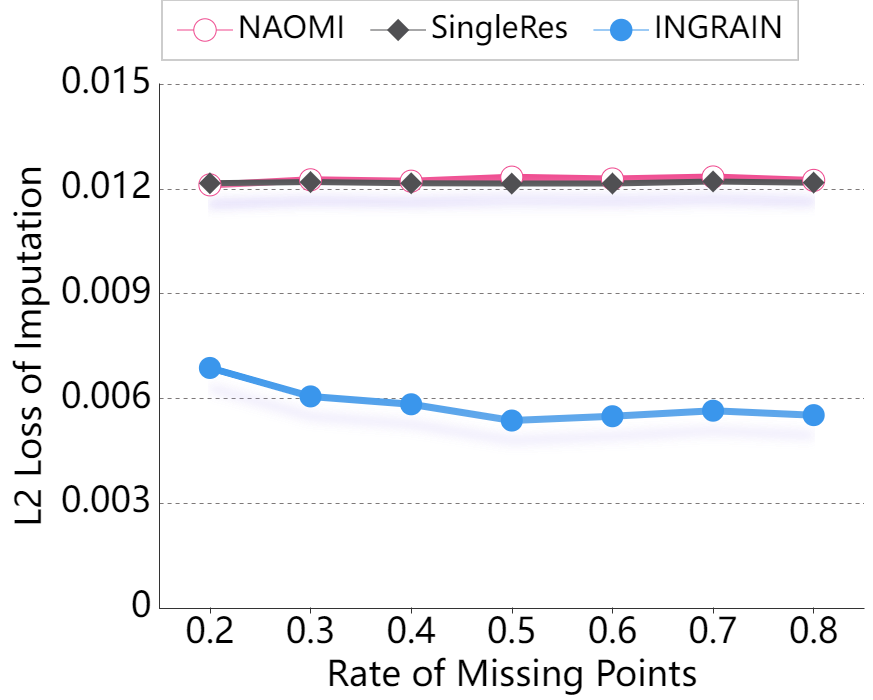}
    } \\
    \subfloat[Geolife - Prediction  \label{subfig:geo_pred_misss_rate}]{%
    \includegraphics[width=0.33\textwidth]{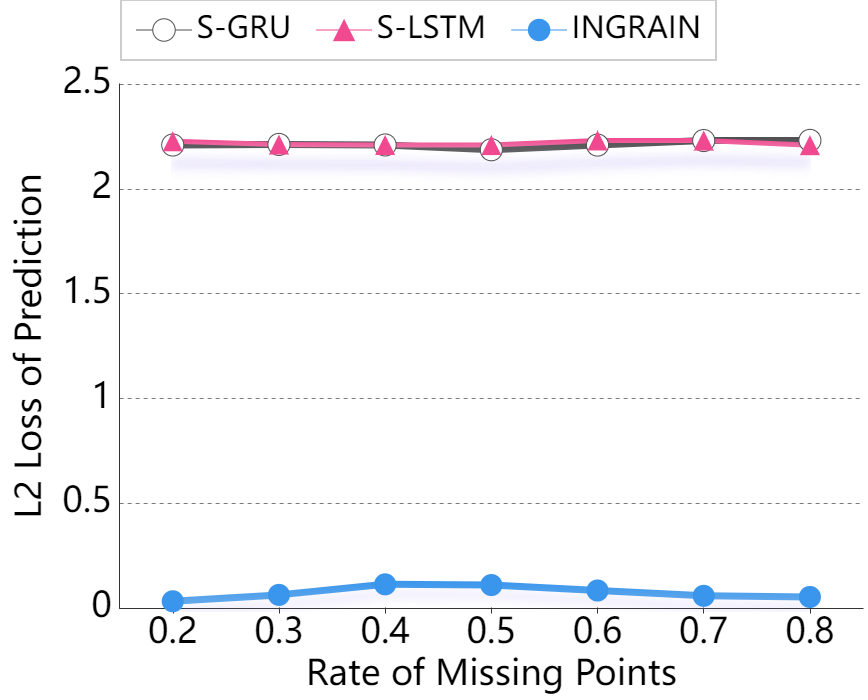}
    }
    \subfloat[Cuebiq-AU - Prediction \label{subfig:cuebiq_au_pred_misss_rate}]{%
    \includegraphics[width=0.33\textwidth]{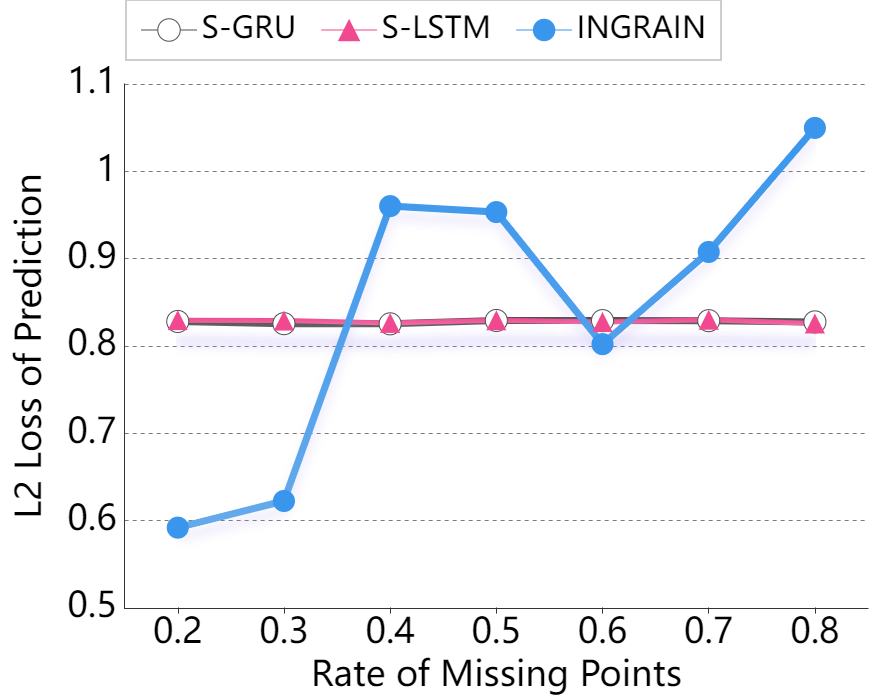}
    }
    \subfloat[Cuebiq-US - Prediction \label{subfig:cuebiq_us_pred_misss_rate}]{%
    \includegraphics[width=0.33\textwidth]{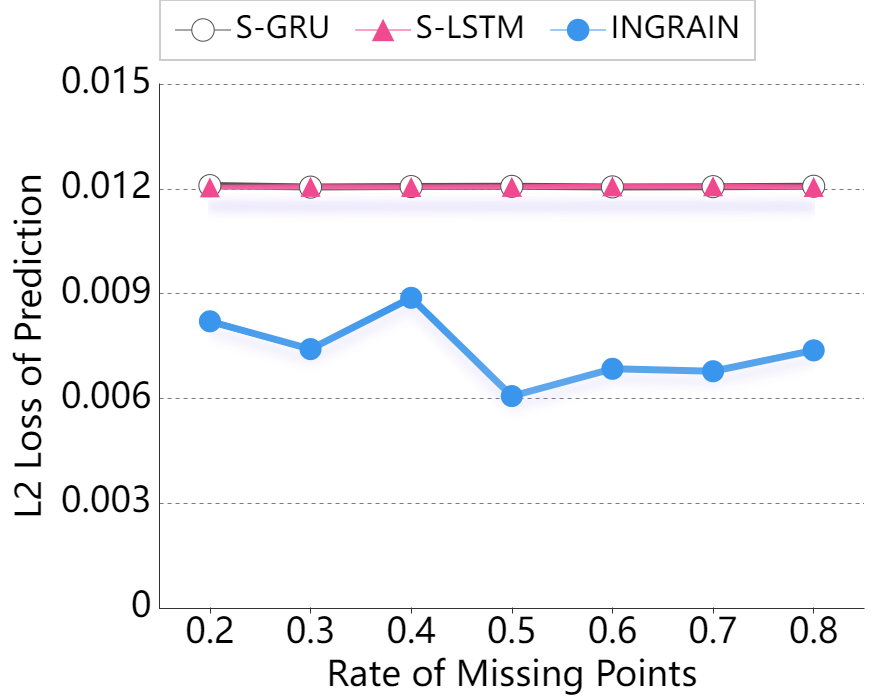}
    }
    \caption{We use the 13 most active users of Geolife and the 20 most active users of Cuebiq-AU and Cuebiq-US for this experiment, and the length of a trajectory is 20. (a), (b) and (c) demonstrate the imputation loss of the proposed model and baselines on different datasets, with varying percentages of missing points in trajectories. (d), (e) and (f) show the results for prediction loss on three datasets, respectively.}
    \label{fig:imp_loss_mis_percent}
\end{figure*}

\subsubsection{Imputation Results}
We first evaluate imputation performance for different lengths of trajectories with a certain degree of missing values. Table \ref{tab:imp_pre_loss} displays the experimental results on three different lengths (20, 50 and 100) of trajectories across all the datasets. The missing rate of points is 0.8. The parameters $\lambda_1$ and $\lambda_2$  are configured to one for our model, which means that the model fully considers the feedback from different components for training. $\lambda_3$ is not considered in this test. Overall, the proposed model has the best imputation performance on these datasets regarding average L2 loss. It is also noticeable that the INGRAIN can keep contributing to a minor loss of imputation when the length of tested trajectories is increased, whereas no such obvious advantage can be found for the baselines. One reason for this could be that longer trajectories contain more sample fragments, and the model can effectively utilize this increment of data to infer missing values. Another argument is that a good combination of attention-based imputation and prediction components can better enable INGRAIN to overcome imputation in longer trajectories. The proposed imputation component is built with an attention mechanism that learns embeddings by weighting the relations between the points in trajectories. And the other two baselines rely on learning embeddings in sequential dependencies of the trajectories, which could be affected more heavily when more random points are missing.

Next, we assess the impact of different missing rates of points on the task of imputation by the algorithms. Fig. \ref{subfig:geo_imp_miss_rate} and \ref{subfig:cuebiq_au_imp_miss_rate} demonstrate the stability of our method in solving imputation on both Geolife and Cuebiq-AU when the percentage of missing points varies from 0.2 to 0.8 with a trajectory length of 20. As the missing rate rises, the INGRAIN can maintain the loss at a relatively low position while the loss of either NAOMI or SingleRes fluctuates dramatically and tends to rise or stay between the missing rate from 0.5 to 0.8. However, the SingleRes performs better on Geolife when the rate is smaller than 0.5. As we can see from Fig. \ref{subfig:cuebiq_us_imp_miss_rate}, the baselines become steadier on the dataset Cuebiq-US, but have (approximately two times) higher loss of imputation than that of our model. In addition, in other tests with different groups of random users, Fig. \ref{fig:imp_loss_group_mis_percent} further demonstrates the model's prominent ability on trajectory imputation in terms of accurate estimation and stability. We claim that learning point embeddings based on the mode of a fully connected graph (attention mechanism) could better capture the dependencies between missing points and the observed trajectories for solving the imputation of daily human mobility in the city regions.

\subsubsection{Prediction Results}
Forward prediction is conducted along with imputation by the proposed model. The results of the proposed model again show its advantages in predicting future values after the imputation of the trajectory is processed. Table \ref{tab:imp_pre_loss} reveals that the INGRAIN is superior to all the other RNN-based baselines (S-GRU, S-LSTM, B-LSTM, and RNNSearch) on both Geolife and Cuebiq-US datasets with missing rates of 0.8. For the dataset Cuebiq-AU, the INGRAIN tends to improve in longer trajectories ($L$=100), although it is worse in shorter ones ($L$=20 or 50) compared with its counterparts. Overall, the results of the baselines deteriorated slightly while the length of trajectories was prolonged, with the same degree of missing rate. Our model shows a more reliable capability to overcome the impact of missing values in longer trajectories for next-location forecasting. 

Moreover, Fig. \ref{subfig:geo_pred_misss_rate}, \ref{subfig:cuebiq_au_pred_misss_rate} and \ref{subfig:cuebiq_us_pred_misss_rate} display the results of prediction with missing rates from 0.2 to 0.8 on three different datasets, respectively. And the length of all the trajectories in this test is 20. The performance of the RNN-based baselines is stable among the three datasets but weaker in regard to the ability to converge at a smaller loss. In contrast, the figure of INGRAIN fluctuates on the Cuebiq-AU but can keep prediction loss at significantly lower values on the other two datasets. Fig. \ref{subfig:cuebiq_au_pred_misss_rate_group} and \ref{subfig:cuebiq_us_pred_misss_rate_group} also show the advantages of the proposed model for prediction tasks with different groups of random users. In general, the previous results indicate that effectively incorporating the imputation component with the prediction unit in INGRAIN would eventually benefit both learning tasks and outperform the counterparts of baselines. We claim that the proposed model conducts the prediction based on the status or effect of imputation on trajectories, which could potentially enhance the performance. 

\subsubsection{Sensitivity Analysis}
\begin{figure*}[!t]
    \centering
    \subfloat[\# of Points per Imp-cycle \label{subfig:geo_eva_imp_num}]{%
    \includegraphics[width=0.32\textwidth]{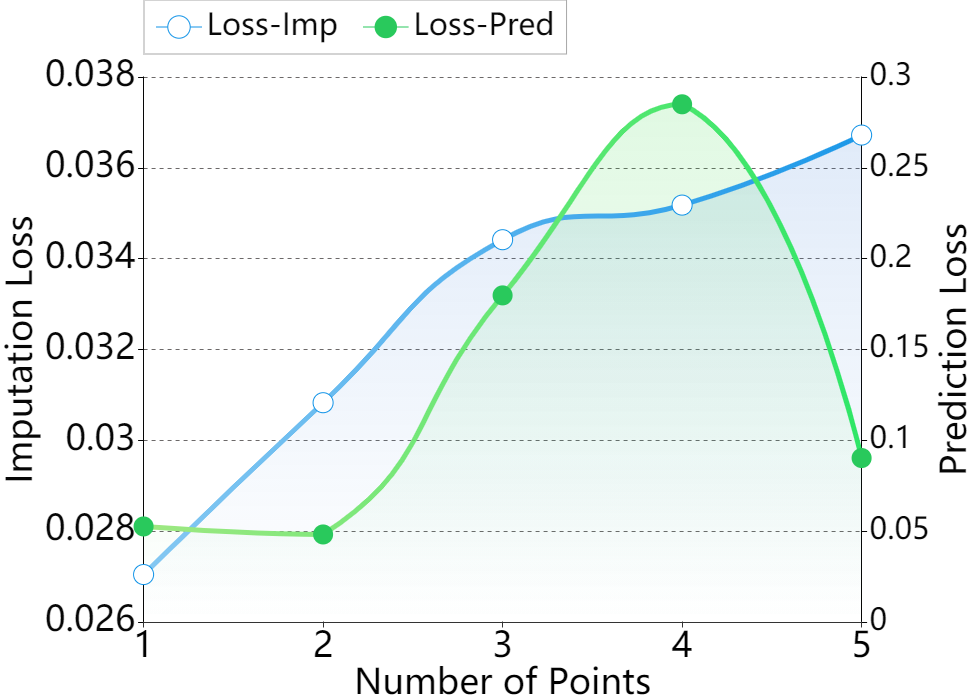}
    } 
    \subfloat[Change of Learning Rate \label{subfig:geo_eva_learn_rate}]{%
    \includegraphics[width=0.32\textwidth]{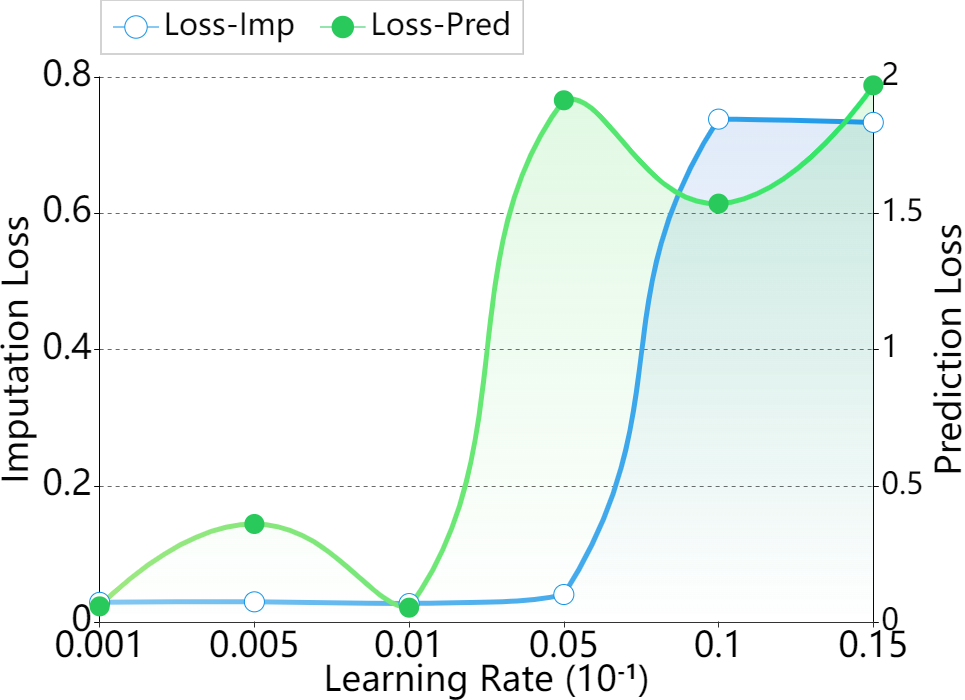}
    }
    \subfloat[Window Length \label{subfig:geo_eva_win_len}]{%
    \includegraphics[width=0.32\textwidth]{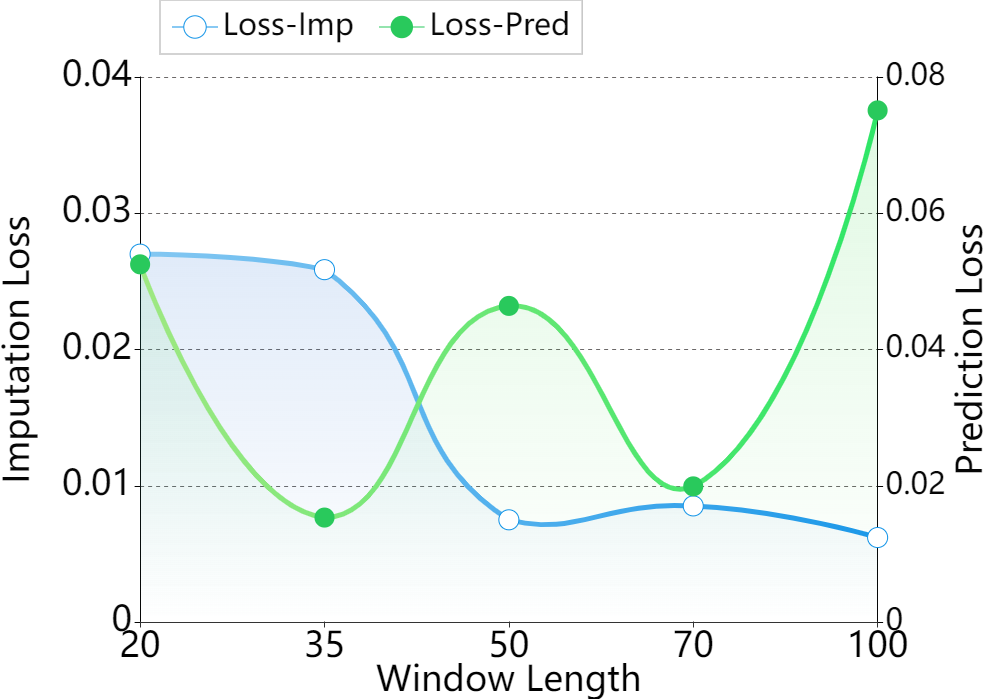}
    }
    \hspace{1pt}
    \subfloat[Head Number \label{subfig:geo_eva_head_num}]{%
    \includegraphics[width=0.32\textwidth]{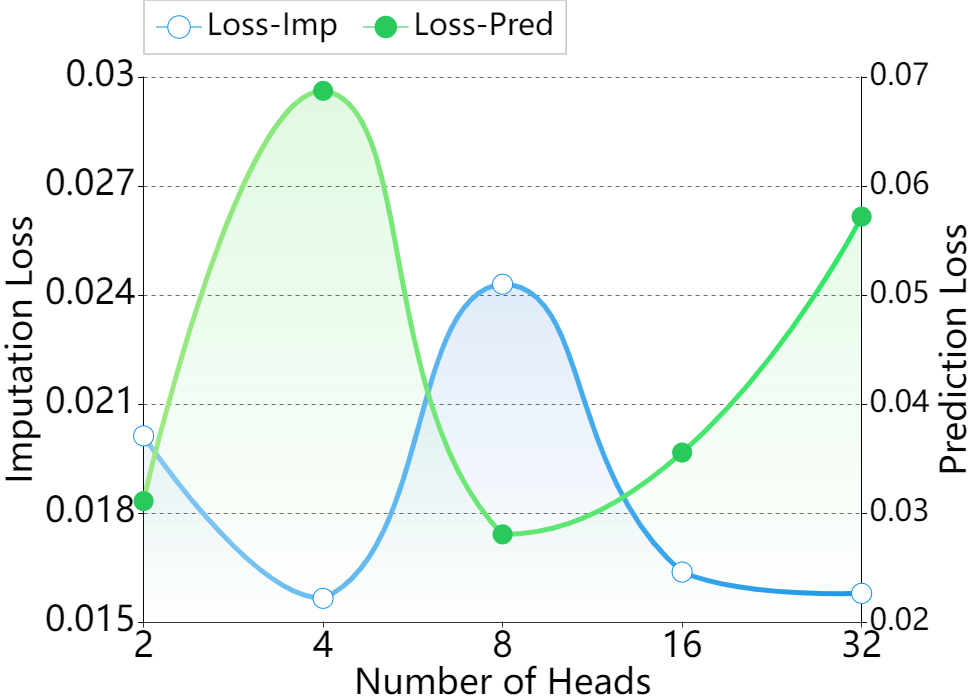}
    }
    \subfloat[Embedding Size \label{subfig:geo_eva_emb_size}]{%
    \includegraphics[width=0.32\textwidth]{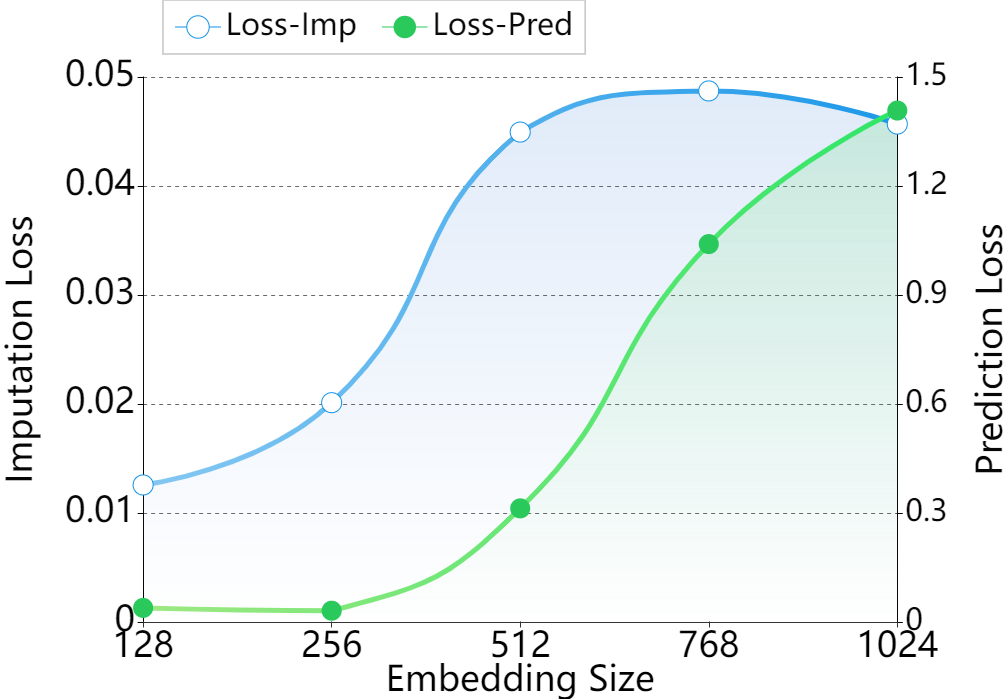}
    }
    \subfloat[Hidden Size \label{subfig:geo_eva_hid_size}]{%
    \includegraphics[width=0.32\textwidth]{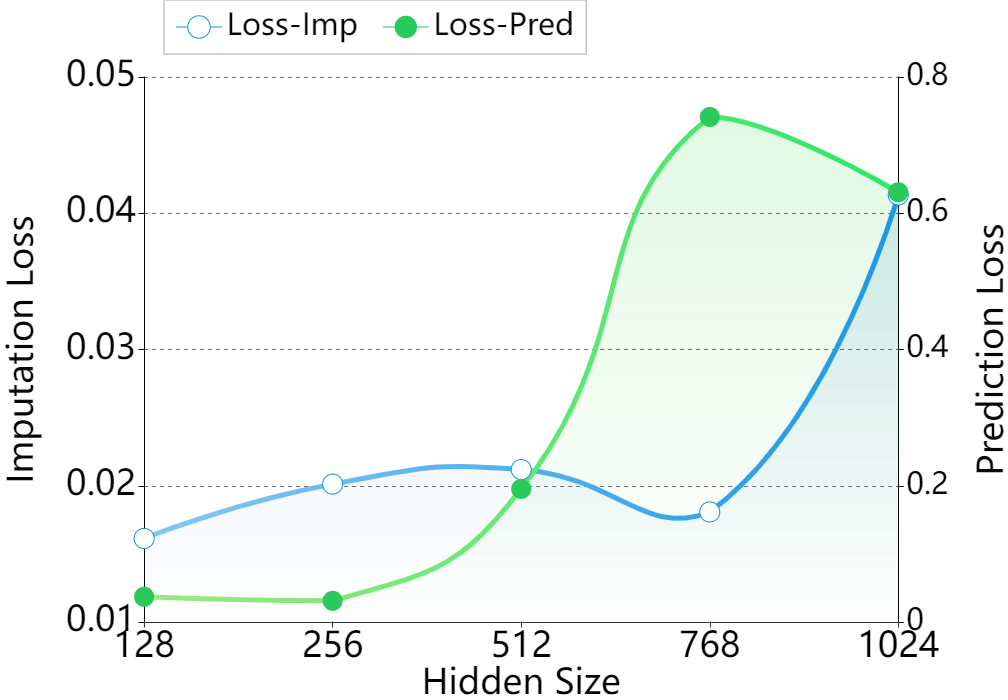}
    }
    \hspace{1pt}
    \subfloat[Evaluation of $\lambda_1$ \label{subfig:geo_eva_para_lamda1}]{%
    \includegraphics[width=0.32\textwidth]{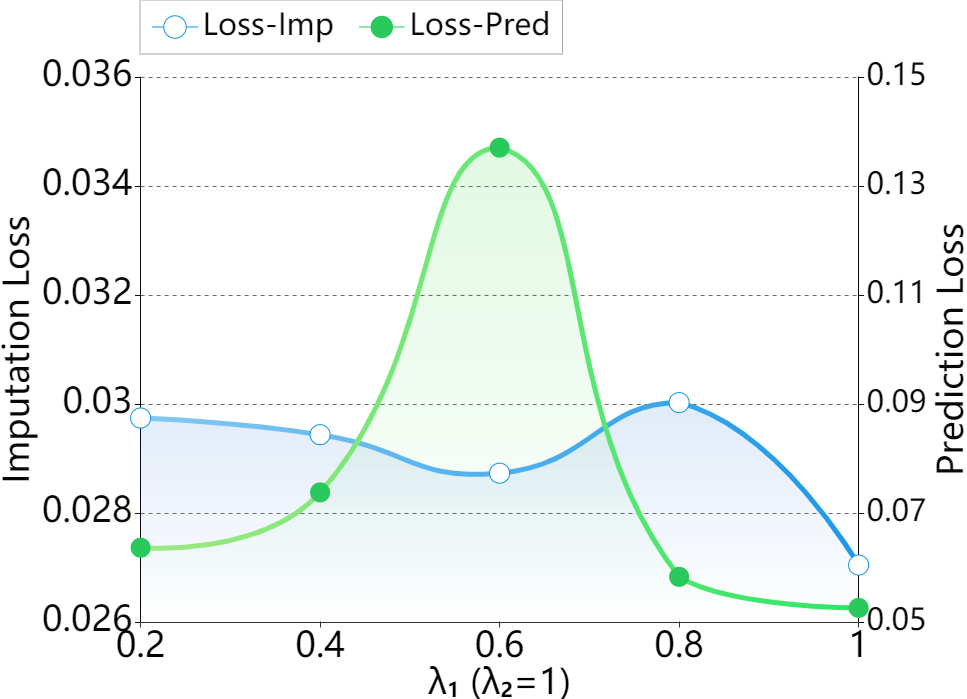}
    }
    \subfloat[Evaluation of $\lambda_2$ \label{subfig:geo_eva_para_lamda2}]{%
    \includegraphics[width=0.32\textwidth]{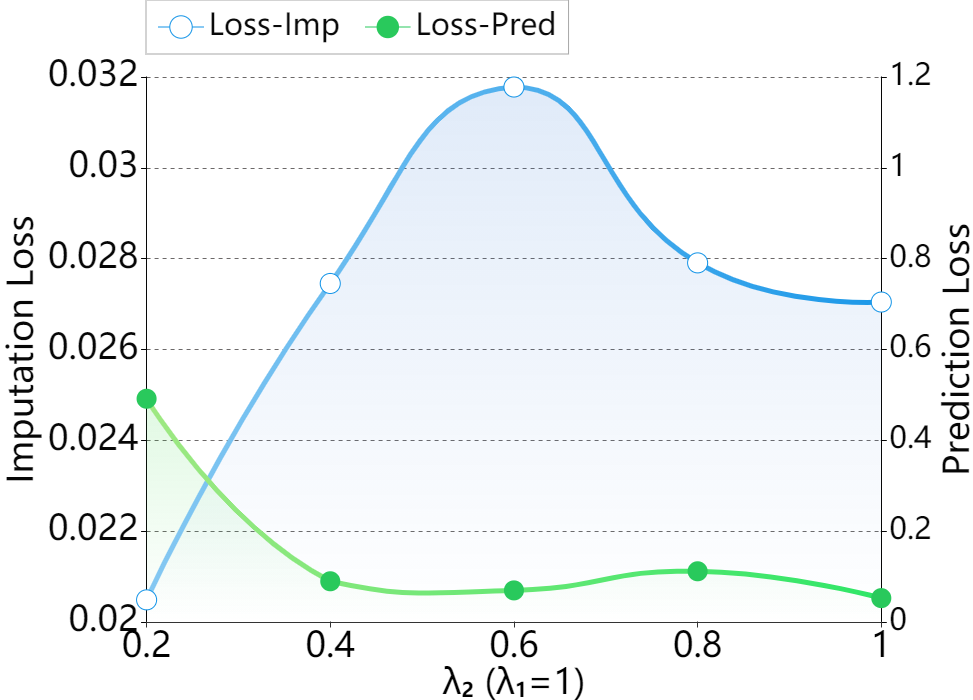}
    }
    \subfloat[Evaluation of $\lambda_3$ \label{subfig:geo_eva_para_lamda3}]{%
    \includegraphics[width=0.32\textwidth]{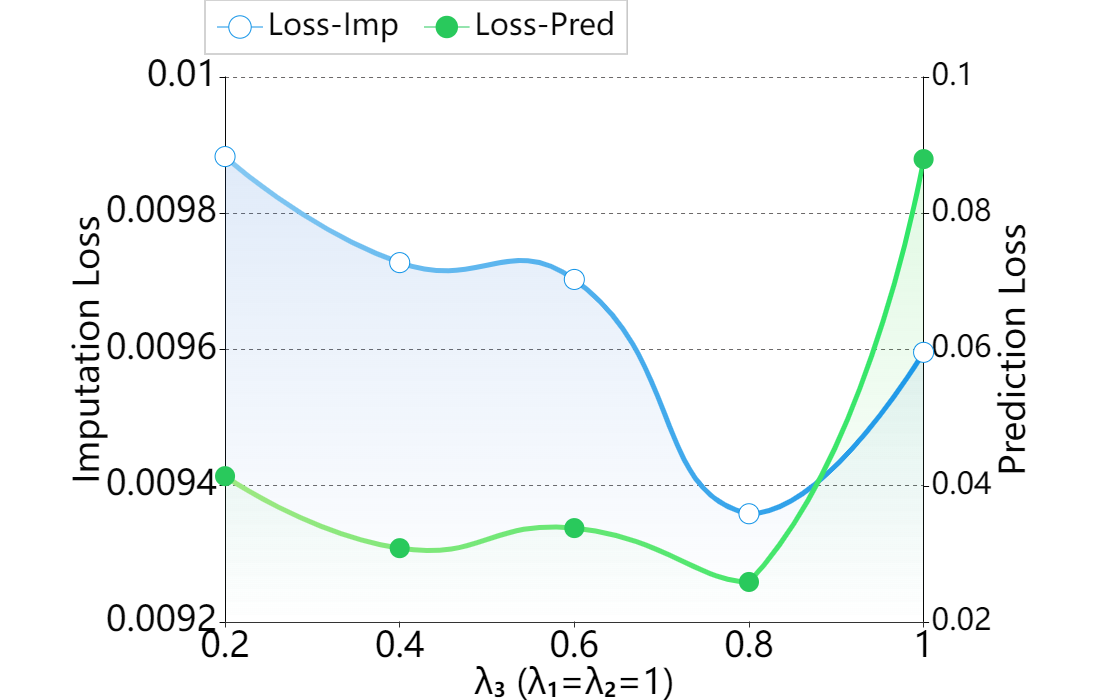}
    }
    \caption{We evaluate the main hyperparameters of the proposed model for both imputation and prediction on Geolife. (a) shows the results for applying different numbers of missing points in each imputing cycle, and (b) illustrates the results with the change in learning rate. (c) and (d) show the evaluation of the model for window length and head number, respectively. The figures related to embedding and hidden size are given in (e) and (f). In addition, (g), (h) and (i) are the results of investigation for $\lambda_1$, $\lambda_2$ and $\lambda_3$, respectively.}
    \label{fig:model_paras_eva}
\end{figure*}

\begin{figure*}[t]
    \centering
    \subfloat[Geolife - Imputation \label{subfig:geo_imp_miss_distri}]{%
    \includegraphics[width=0.33\textwidth]{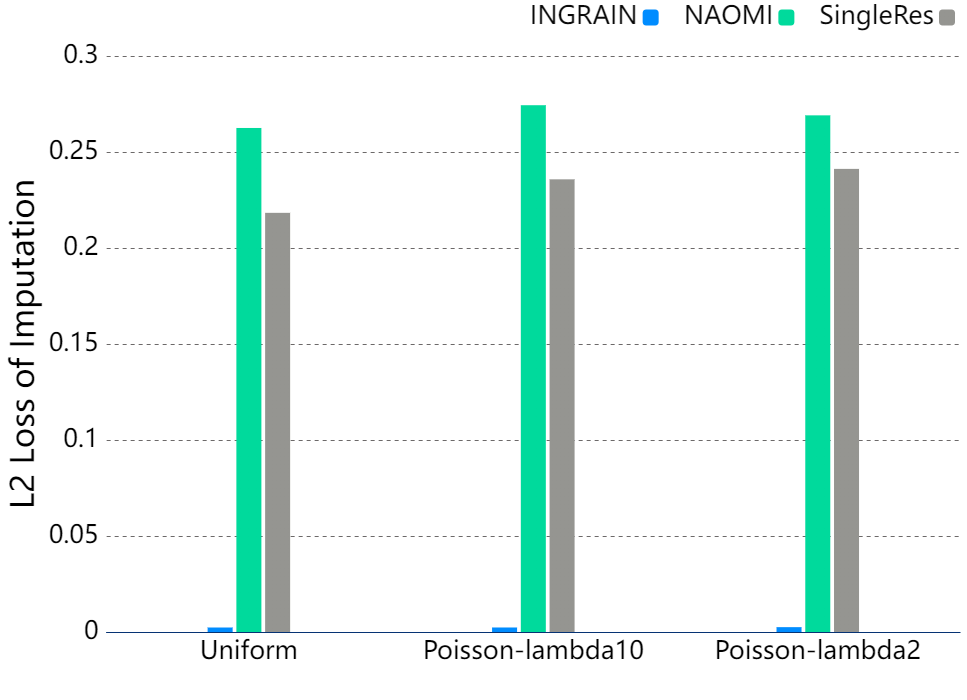}
    }
    \subfloat[Cuebiq-AU - Imputation \label{subfig:cuebiq_au_imp_miss_distri}]{%
    \includegraphics[width=0.33\textwidth]{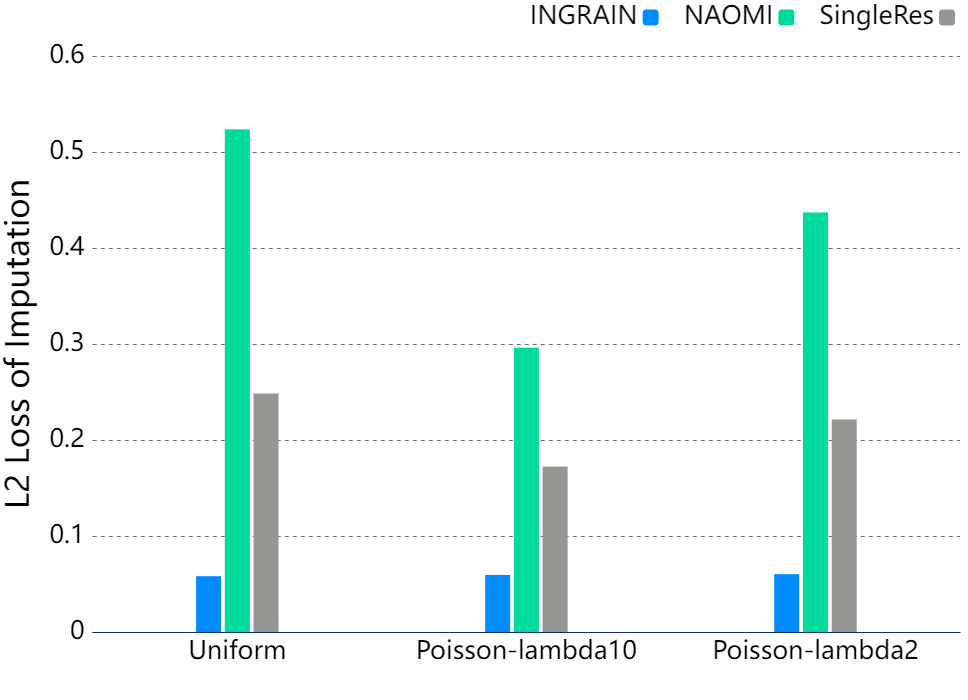}
    }
    \subfloat[Cuebiq-US - Imputation \label{subfig:cuebiq_us_imp_miss_distri}]{%
    \includegraphics[width=0.33\textwidth]{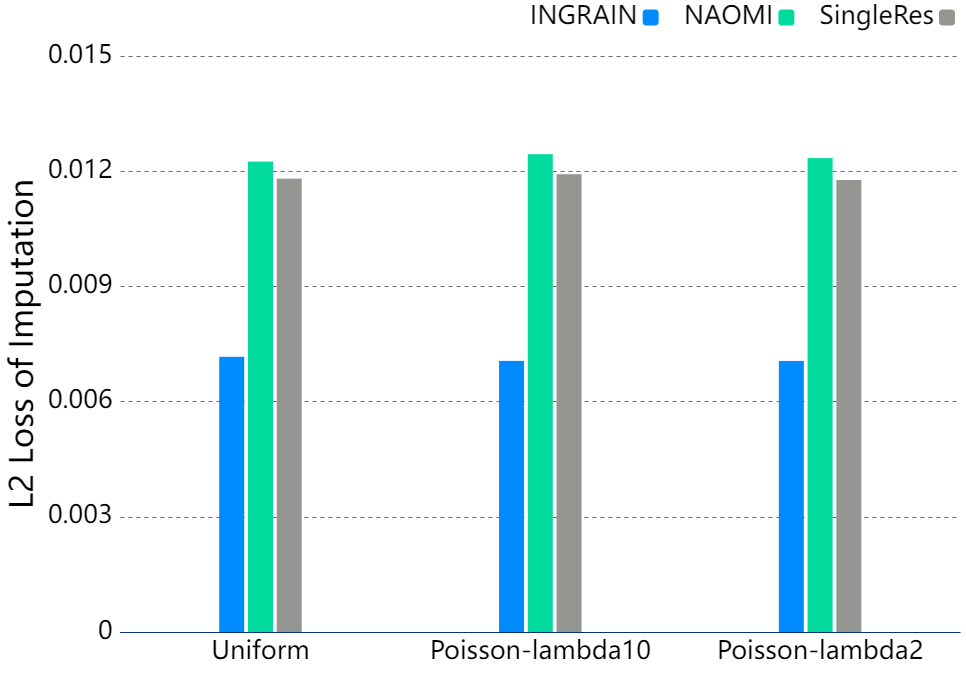}
    } \\
    \subfloat[Geolife - Prediction \label{subfig:geo_pred_miss_distri}]{%
    \includegraphics[width=0.33\textwidth]{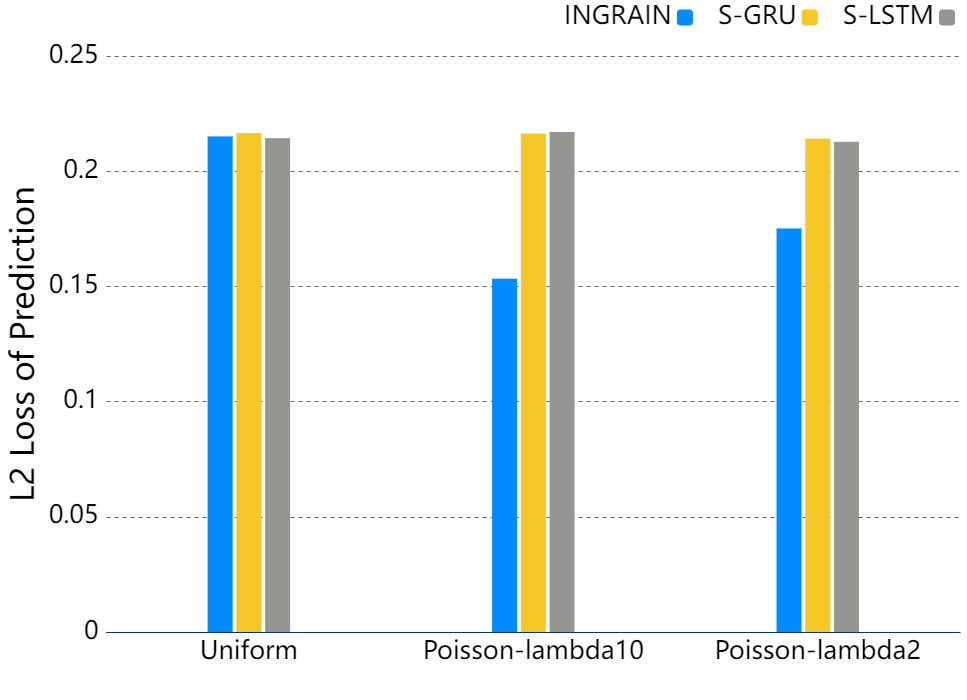}
    }
    \subfloat[Cuebiq-AU - Prediction \label{subfig:cuebiq_au_pred_miss_distri}]{%
    \includegraphics[width=0.33\textwidth]{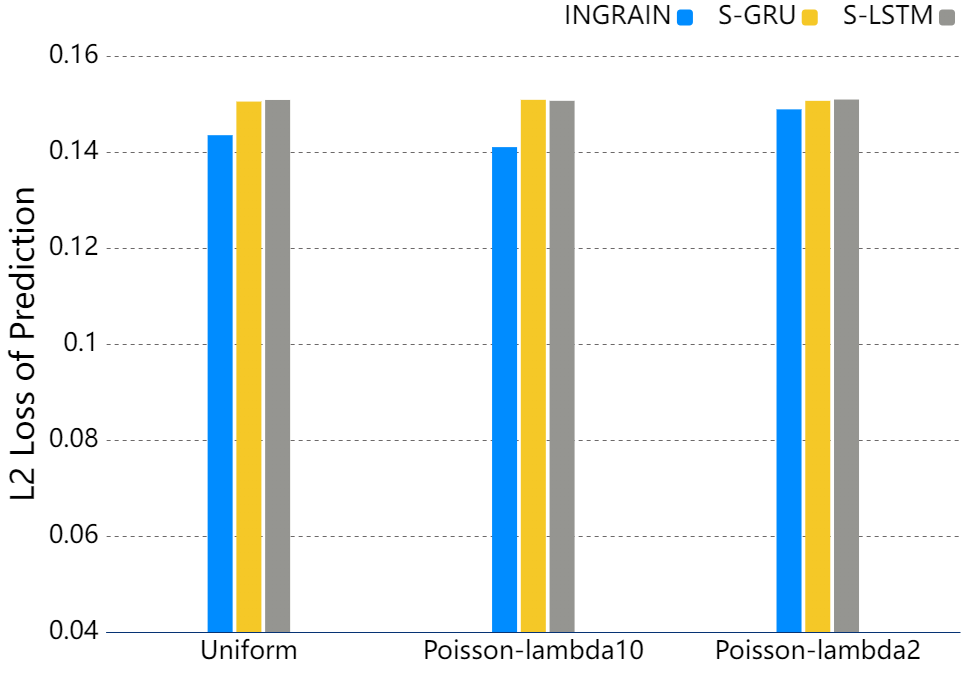}
    }
    \subfloat[Cuebiq-US - Prediction \label{subfig:cuebiq_us_pred_miss_distri}]{%
    \includegraphics[width=0.33\textwidth]{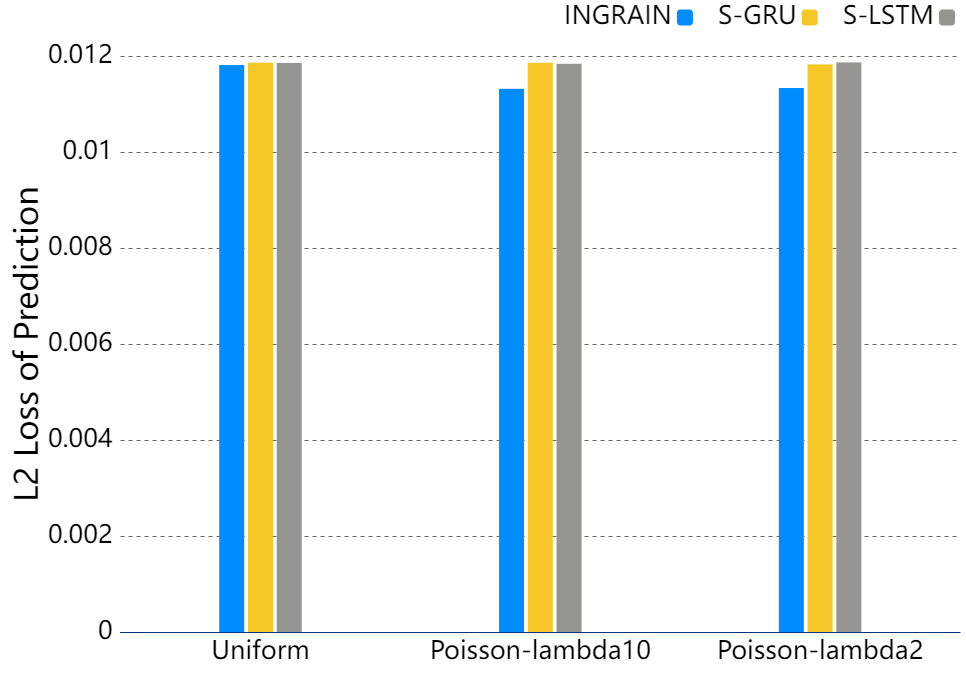}
    }
    \caption{This experiment was run with a group of 10 users randomly picked from the 30 most active ones in each dataset, and the trajectory length is 20. (a), (b) and (c) demonstrate the imputation loss of the proposed model and baselines on three datasets, with different distributions of missing values generation. (d), (e) and (f) show the results for the prediction task on three datasets, respectively.}
    \label{fig:model_miss_distri_eva}
\end{figure*}

In this section, we examine the effect of primary hyperparameters on the performance of INGRAIN for both tasks. That gives us more insights into how the proposed method converges in different configurations and what trade-offs can be made between two learning tasks. As this paper mentioned, the model is agile to infer a defined number of missing points in each imputing cycle. This process iterates until the whole imputation work of each trajectory is finished. Thus, we first check how this number potentially acts on the results of two learning tasks with the dataset Geolife. In Fig. \ref{subfig:geo_eva_imp_num}, the number of missing points from 1 to 5 per imputing cycle is inspected for the best average L2 loss of imputation and prediction in each test simultaneously. There is an increase in the loss for both criteria when more points are imputed in each cycle. The loss of imputation and prediction is relatively minor when the number is one. Although the prediction loss arrives at its lowest position when the number is two, the imputation loss value soars. Therefore, using fewer missing points in each imputation operation can offer better learning results for both tasks. It is well known that the learning rate is a fundamental factor that affects the convergence of a deep learning model. Fig. \ref{subfig:geo_eva_learn_rate} illustrates that the performance of two tasks becomes gradually worse when we increase the value of the learning rate from 0.001 ($10^{-1}$) to 0.15 ($10^{-1}$). As we can see, learning rates 0.001 ($10^{-1}$) and 0.01 ($10^{-1}$) allow the model to produce better results for both imputation and prediction.

In the assessment of using a different window length for constructing trajectories, Fig. \ref{subfig:geo_eva_win_len} shows that imputation loss tends to become smaller in the learning with longer trajectories, and prediction loss fluctuates slightly between around 0.02 to 0.07. We see that the performance of both imputation and prediction work is relatively advantageous when the length is 70. Fig. \ref{subfig:geo_eva_head_num} indicates that using two heads of self-attention can simultaneously perform better for both learning tasks. Furthermore, the computational requirement becomes relatively less than applying a bigger number of heads. In addition, the evaluation of embedding and hidden size used in the model demonstrates that a smaller value can already contribute to a good performance of two tasks, such as 128 or 256. As Fig. \ref{subfig:geo_eva_emb_size} and Fig. \ref{subfig:geo_eva_hid_size} show, more computation with an increased value of such parameters did not provide better results.

In the following, we examine the functions of imputation and prediction components by changing the values of $\lambda_1$ and $\lambda_2$ without considering the constraint of movement velocity ($\lambda_3=0$). As we mentioned in Section \ref{sec:collaborative_learning_objective}, $\lambda_1$ and $\lambda_2$ are two hyperparameters that control the feedback of imputation and prediction unit received by the model during training, respectively. In Fig. \ref{subfig:geo_eva_para_lamda1} and \ref{subfig:geo_eva_para_lamda2}, higher values indicate that more strength is considered for the corresponding component during entire training. We see that the loss of imputation reduces slightly in Fig. \ref{subfig:geo_eva_para_lamda1} when bigger $\lambda_1$ is configured. Meanwhile, the prediction loss simultaneously undergoes a reversed change (rise). However, the combination $\lambda_1=1$ and $\lambda_2=1$ could offer a better trade-off for the performance of both tasks. Similarly, in a different run, the loss of prediction decreases drastically while the value of $\lambda_2$ is raised from 0.2 to 1 along with a fixed $\lambda_1$ in Fig. \ref{subfig:geo_eva_para_lamda2}, and both tasks are beneficial when $\lambda_2$ is round 0.3. In addition, Fig. \ref{subfig:geo_eva_para_lamda3} presents the results of varied $\lambda_3$ when $\lambda_1=1$ and $\lambda_2=1$. It is apparent that the performance of both task benefit at most when $\lambda_3=0.8$.

In Fig. \ref{fig:model_miss_distri_eva}, we compare the performance of the proposed model and the baselines with different distributions of missing values generation, such as Uniform and Poisson distribution. Fig. \ref{subfig:geo_imp_miss_distri}, \ref{subfig:cuebiq_au_imp_miss_distri} and \ref{subfig:cuebiq_us_imp_miss_distri} demonstrate the imputation loss of the proposed model and baselines on different datasets, respectively. Fig. \ref{subfig:geo_pred_miss_distri}, \ref{subfig:cuebiq_au_pred_miss_distri} and \ref{subfig:cuebiq_us_pred_miss_distri} are the results for the prediction task on three datasets, respectively. As we can see, when using various distributions for the experiment, the proposed approach exhibits significant advantages on the imputation task. Meanwhile, our model can still keep a competitive performance on the prediction task compared with the other algorithms.

\subsubsection{Ablation Study}
\begingroup
\setlength{\tabcolsep}{8pt}
\begin{table}[t]
\centering
\caption{The model will not consider the feedback from the imputation unit during training when $\lambda_1$=0. Otherwise, $\lambda_1$=1. $\lambda_2$ and $\lambda_3$ are responsible for controlling the feedback from the prediction unit and the weight of speed constraint, respectively. \textit{Loss-I} and \textit{Loss-P} represent the L2 loss of imputation and prediction, respectively. Further, the portion of missing points is 0.8 for all the tests, and \textit{L} denotes the length of trajectories in different trials. Basically, the learning of a task is infeasible if its designated optimization is totally ignored. However, another task has the chance of getting a slight improvement than considering more optimization units in the same iterations of training.}
\label{tab:ablation_lamda}
\begin{tabular}{ccc|c|cc|cc|cc} 
\hline\hline
\multicolumn{3}{c|}{\begin{tabular}[c]{@{}c@{}}\\\textbf{\textbf{\textbf{\textbf{Weights}}}}\\\textbf{\textbf{\textbf{\textbf{}}}}\end{tabular}} & \multirow{2}{*}{\begin{tabular}[c]{@{}c@{}}\textbf{\textbf{Tasks}}\\\end{tabular}} & \multicolumn{2}{c|}{\textbf{Geolife }}                       & \multicolumn{2}{c|}{\textbf{Cuebiq - AU}}                    & \multicolumn{2}{c}{\textbf{\textbf{Cuebiq - US}}}  \\ 
\cline{1-3}\cline{5-10}
\multicolumn{1}{l}{\textit{$\lambda_1$}} & \multicolumn{1}{l}{\textit{$\lambda_2$}} & \multicolumn{1}{l|}{\textit{$\lambda_3$}}                  &                                                                                    & \textit{L} = 20                            & \textit{L} = 50 & \textit{L} = 20 & \textit{L} = 50                            & \textit{L} = 20 & \textit{L} = 50                  \\ 
\hline
\multirow{2}{*}{1}                       & \multirow{2}{*}{1}                       & \multirow{2}{*}{0}                                         & \textit{Loss-I}                                                                    & 0.0270 & 0.0075          & 0.0122          & 0.0117                                     & 0.0055          & 0.0050                           \\
                                         &                                          &                                                            & \textit{Loss-P}                                                                    & 0.0525                                     & 0.0464          & 1.0497          & 1.0140                                     & 0.0073          & 0.0070                           \\ 
\hline
\multirow{2}{*}{1}                       & \multirow{2}{*}{0}                       & \multirow{2}{*}{0}                                         & \textit{Loss-I}                                                                    & 0.0275                                     & 0.0098          & 0.0113          & 0.0109                                     & 0.0054          & 0.0055                           \\
                                         &                                          &                                                            & \textit{Loss-P}                                                                    & $\sim$                                     & $\sim$          & $\sim$          & $\sim$                                     & $\sim$          & $\sim$                           \\ 
\hline
\multirow{2}{*}{0}                       & \multirow{2}{*}{1}                       & \multirow{2}{*}{0 }                                        & \textit{Loss-I}                                                                    & $\sim$                                     & $\sim$          & $\sim$          & $\sim$                                     & $\sim$          & $\sim$                           \\
                                         &                                          &                                                            & \textit{Loss-P}                                                                    & 0.0959                                     & 0.0393          & 0.8611          & 1.1230                                     & 0.0114          & 0.0082                           \\ 
\hline
\multirow{2}{*}{1}                       & \multirow{2}{*}{1}                       & \multirow{2}{*}{1}                                         & \multicolumn{1}{l|}{\textit{Loss-I}}                                               & 0.0107                                     & 0.0062          & 0.0120          & 0.0110 & 0.0055          & 0.0048                           \\
                                         &                                          &                                                            & \multicolumn{1}{l|}{\textit{Loss-P}}                                               & 0.0648                                     & 0.0499          & 0.9350          & 1.2609                                     & 0.0070          & 0.0076                           \\
\hline
\end{tabular}
\end{table}
\endgroup

\begingroup
\setlength{\tabcolsep}{10pt}
\begin{table}[t]
\centering
\caption{The RNN-based unit and the Supplement layer are two modules that support the imputation and prediction learning process. The table shows the impact of these two modules on the model's performance. 'Add operation' or 'Replace operation' is used individually in the Supplement layer with or without the RNN unit. The tests were conducted on Geolife with a trajectory length of 20.}
\label{tab:ablation_operations}
\begin{tabular}{ccc|cc} 
\hline\hline
\multicolumn{3}{c|}{\textbf{Components}}                                                      & \multicolumn{2}{c}{\textbf{Geolife}}  \\ 
\hline
\textit{RNN Unit}            & \textit{Add Operation}        & \textit{Replace Operation}    & \textit{Loss-I} & \textit{Loss-P}     \\ 
\hline
-                             & -                             & -                             & 0.0105          & 0.0143              \\ 
\hline
$\checkmark$ & -                             & -                             & 0.0097          & 0.0480              \\ 
\hline
-                             & -                             & $\checkmark$ & 0.0159          & 0.0695              \\ 
\hline
$\checkmark$ & $\checkmark$ & -                             & 0.0094          & 0.0732              \\ 
\hline
$\checkmark$ & -                             & $\checkmark$ & 0.0090          & 0.0886              \\
\hline
\end{tabular}
\end{table}
\endgroup

We conduct an ablation study of $\lambda_1$, $\lambda_2$, and $\lambda_3$ to check the model's performance when feedback of the imputation component, prediction component, or speed constraint is totally discarded during optimization. The proposed model will fully consider the feedback of imputation in the optimization process when $\lambda_1=1$ and $\lambda_2=1$ indicates that the optimizer will receive the feedback of the prediction component without any abandon. In contrast, zero value means that the feedback from one component is totally left out during training. It is found that when the feedback from one task is totally discarded during the training of the whole model, the outputs of that task will be meaningless and random floats because no specific optimization arises based on the ground truth. Thus, we omit those relevant results on the Table \ref{tab:ablation_lamda}. We temporarily neglect the movement speed constraint by setting $\lambda_3$ to zero in the beginning. Furthermore, the table illustrates that the setting of $\lambda_1=1$ and $\lambda_2=0$ could achieve better imputation results in some cases, accompanied by the sacrifice of optimization for the prediction. However, considering the full feedback of both components ($\lambda_1=1, \lambda_2=1$) could also provide competitive prediction performance in most cases (the length of trajectory in 20 or 50). This experiment also indicates that switching one of two main components could give the model the flexibility to concentrate better on optimizing a single task. In addition, adding movement speed constraint ($\lambda_3=1$) could lead to a slight improvement of imputation on Geolife, which has a more stable sampling rate of points collection than the other two datasets.

As mentioned before, the RNN-based unit and the Supplement layer are two modules that support the imputation and prediction learning process. An evaluation is given of their effects on the overall performance of the proposed model in Table \ref{tab:ablation_operations}. 'Add operation' or 'Replace operation' is used individually in the Supplement layer to incorporate missing values' embedding into the trajectory representation. Five tests were conducted for each combination on Geolife with a trajectory length of 20, and the mean values were reported. We can see that combining a supplement operation and an RNN unit tends to contribute to a minor loss of imputation. In contrast, the prediction can not benefit too much from the joint use of two components. However, we can find that integration of RNN unit for learning can improve imputation performance to some extent.

\section{Conclusion}
\label{sec:conclusion}
Usually, human mobility data are incomplete in practice, leading to bias or difficulties in learning tasks, such as imputation and prediction. We propose a new approach incorporating non-autoregressive and autoregressive components to help trajectory imputation and prediction. The model effectively learns the dependence between observations and missing values on multiple levels with the advantage of self-attention. Meanwhile, one RNN-based unit is applied to extract potential features recurrently from the newly learned sequences. Intensive experiments are conducted on three datasets: Geolife, Cuebiq-AU, and Cuebiq-US. The results show that the proposed model can achieve advanced performance in both learning tasks compared to the baselines. Additionally, the analysis of primary hyperparameters reveals how trade-offs could be made between different tasks with proper settings. Moreover, the flexible configuration of switching the acceptance of additional feedback enables us to pay more attention to individual units to attain better results for a specific task. In the future, we plan to conduct more experiments on more diverse types of mobility datasets (e.g., POIs and grid-based datasets) and analyze the potential factors that crucially influence the learning of different imputation algorithms.


\bibliographystyle{ACM-Reference-Format}
\bibliography{references}


\begin{thebibliography}{38}


\ifx \showCODEN    \undefined \def \showCODEN     #1{\unskip}     \fi
\ifx \showDOI      \undefined \def \showDOI       #1{#1}\fi
\ifx \showISBNx    \undefined \def \showISBNx     #1{\unskip}     \fi
\ifx \showISBNxiii \undefined \def \showISBNxiii  #1{\unskip}     \fi
\ifx \showISSN     \undefined \def \showISSN      #1{\unskip}     \fi
\ifx \showLCCN     \undefined \def \showLCCN      #1{\unskip}     \fi
\ifx \shownote     \undefined \def \shownote      #1{#1}          \fi
\ifx \showarticletitle \undefined \def \showarticletitle #1{#1}   \fi
\ifx \showURL      \undefined \def \showURL       {\relax}        \fi
\providecommand\bibfield[2]{#2}
\providecommand\bibinfo[2]{#2}
\providecommand\natexlab[1]{#1}
\providecommand\showeprint[2][]{arXiv:#2}

\bibitem[\protect\citeauthoryear{Alahi, Goel, Ramanathan, Robicquet, Fei-Fei,
  and Savarese}{Alahi et~al\mbox{.}}{2016}]%
        {alahi2016social}
\bibfield{author}{\bibinfo{person}{Alexandre Alahi}, \bibinfo{person}{Kratarth
  Goel}, \bibinfo{person}{Vignesh Ramanathan}, \bibinfo{person}{Alexandre
  Robicquet}, \bibinfo{person}{Li Fei-Fei}, {and} \bibinfo{person}{Silvio
  Savarese}.} \bibinfo{year}{2016}\natexlab{}.
\newblock \showarticletitle{Social lstm: Human trajectory prediction in crowded
  spaces}. In \bibinfo{booktitle}{\emph{IEEE conference on computer vision and
  pattern recognition}}. \bibinfo{pages}{961--971}.
\newblock


\bibitem[\protect\citeauthoryear{Bahdanau, Cho, and Bengio}{Bahdanau
  et~al\mbox{.}}{2014}]%
        {bahdanau2014neural}
\bibfield{author}{\bibinfo{person}{Dzmitry Bahdanau},
  \bibinfo{person}{Kyunghyun Cho}, {and} \bibinfo{person}{Yoshua Bengio}.}
  \bibinfo{year}{2014}\natexlab{}.
\newblock \showarticletitle{Neural machine translation by jointly learning to
  align and translate}.
\newblock \bibinfo{journal}{\emph{arXiv preprint arXiv:1409.0473}}
  (\bibinfo{year}{2014}).
\newblock


\bibitem[\protect\citeauthoryear{Cao, Wang, Li, Zhou, Li, and Li}{Cao
  et~al\mbox{.}}{2018}]%
        {cao2018brits}
\bibfield{author}{\bibinfo{person}{Wei Cao}, \bibinfo{person}{Dong Wang},
  \bibinfo{person}{Jian Li}, \bibinfo{person}{Hao Zhou}, \bibinfo{person}{Lei
  Li}, {and} \bibinfo{person}{Yitan Li}.} \bibinfo{year}{2018}\natexlab{}.
\newblock \showarticletitle{Brits: Bidirectional recurrent imputation for time
  series}. In \bibinfo{booktitle}{\emph{Advances in Neural Information
  Processing Systems}}. \bibinfo{pages}{6775--6785}.
\newblock


\bibitem[\protect\citeauthoryear{Chen, Ma, Susilo, Liu, and Wang}{Chen
  et~al\mbox{.}}{2016}]%
        {chen2016promises}
\bibfield{author}{\bibinfo{person}{Cynthia Chen}, \bibinfo{person}{Jingtao Ma},
  \bibinfo{person}{Yusak Susilo}, \bibinfo{person}{Yu Liu}, {and}
  \bibinfo{person}{Menglin Wang}.} \bibinfo{year}{2016}\natexlab{}.
\newblock \showarticletitle{The promises of big data and small data for travel
  behavior (aka human mobility) analysis}.
\newblock \bibinfo{journal}{\emph{Transportation research part C: emerging
  technologies}}  \bibinfo{volume}{68} (\bibinfo{year}{2016}),
  \bibinfo{pages}{285--299}.
\newblock


\bibitem[\protect\citeauthoryear{Chung, Gulcehre, Cho, and Bengio}{Chung
  et~al\mbox{.}}{2014}]%
        {chung2014empirical}
\bibfield{author}{\bibinfo{person}{Junyoung Chung}, \bibinfo{person}{Caglar
  Gulcehre}, \bibinfo{person}{KyungHyun Cho}, {and} \bibinfo{person}{Yoshua
  Bengio}.} \bibinfo{year}{2014}\natexlab{}.
\newblock \showarticletitle{Empirical evaluation of gated recurrent neural
  networks on sequence modeling}.
\newblock \bibinfo{journal}{\emph{arXiv preprint arXiv:1412.3555}}
  (\bibinfo{year}{2014}).
\newblock


\bibitem[\protect\citeauthoryear{Fedus, Goodfellow, and Dai}{Fedus
  et~al\mbox{.}}{2018}]%
        {fedus2018maskgan}
\bibfield{author}{\bibinfo{person}{William Fedus}, \bibinfo{person}{Ian
  Goodfellow}, {and} \bibinfo{person}{Andrew~M Dai}.}
  \bibinfo{year}{2018}\natexlab{}.
\newblock \showarticletitle{MaskGAN: Better text generation via filling in
  the\_}.
\newblock \bibinfo{journal}{\emph{arXiv preprint arXiv:1801.07736}}
  (\bibinfo{year}{2018}).
\newblock


\bibitem[\protect\citeauthoryear{Feng, Li, Zhang, Sun, Meng, Guo, and Jin}{Feng
  et~al\mbox{.}}{2018}]%
        {feng2018deepmove}
\bibfield{author}{\bibinfo{person}{Jie Feng}, \bibinfo{person}{Yong Li},
  \bibinfo{person}{Chao Zhang}, \bibinfo{person}{Funing Sun},
  \bibinfo{person}{Fanchao Meng}, \bibinfo{person}{Ang Guo}, {and}
  \bibinfo{person}{Depeng Jin}.} \bibinfo{year}{2018}\natexlab{}.
\newblock \showarticletitle{Deepmove: Predicting human mobility with
  attentional recurrent networks}. In \bibinfo{booktitle}{\emph{2018 world wide
  web conference}}. \bibinfo{pages}{1459--1468}.
\newblock


\bibitem[\protect\citeauthoryear{Gid{\'o}falvi and Dong}{Gid{\'o}falvi and
  Dong}{2012}]%
        {gidofalvi2012and}
\bibfield{author}{\bibinfo{person}{Gy{\H{o}}z{\H{o}} Gid{\'o}falvi} {and}
  \bibinfo{person}{Fang Dong}.} \bibinfo{year}{2012}\natexlab{}.
\newblock \showarticletitle{When and where next: individual mobility
  prediction}. In \bibinfo{booktitle}{\emph{SIGSPATIAL}}.
  \bibinfo{pages}{57--64}.
\newblock


\bibitem[\protect\citeauthoryear{Giuliari, Hasan, Cristani, and
  Galasso}{Giuliari et~al\mbox{.}}{2020}]%
        {giuliari2020transformer}
\bibfield{author}{\bibinfo{person}{Francesco Giuliari}, \bibinfo{person}{Irtiza
  Hasan}, \bibinfo{person}{Marco Cristani}, {and} \bibinfo{person}{Fabio
  Galasso}.} \bibinfo{year}{2020}\natexlab{}.
\newblock \showarticletitle{Transformer Networks for Trajectory Forecasting}.
\newblock \bibinfo{journal}{\emph{arXiv preprint arXiv:2003.08111}}
  (\bibinfo{year}{2020}).
\newblock


\bibitem[\protect\citeauthoryear{Graves, Fern{\'a}ndez, and Schmidhuber}{Graves
  et~al\mbox{.}}{2005}]%
        {graves2005bidirectional}
\bibfield{author}{\bibinfo{person}{Alex Graves}, \bibinfo{person}{Santiago
  Fern{\'a}ndez}, {and} \bibinfo{person}{J{\"u}rgen Schmidhuber}.}
  \bibinfo{year}{2005}\natexlab{}.
\newblock \showarticletitle{Bidirectional LSTM networks for improved phoneme
  classification and recognition}. In \bibinfo{booktitle}{\emph{International
  conference on artificial neural networks}}. Springer,
  \bibinfo{pages}{799--804}.
\newblock


\bibitem[\protect\citeauthoryear{Gupta, Johnson, Fei-Fei, Savarese, and
  Alahi}{Gupta et~al\mbox{.}}{2018}]%
        {gupta2018social}
\bibfield{author}{\bibinfo{person}{Agrim Gupta}, \bibinfo{person}{Justin
  Johnson}, \bibinfo{person}{Li Fei-Fei}, \bibinfo{person}{Silvio Savarese},
  {and} \bibinfo{person}{Alexandre Alahi}.} \bibinfo{year}{2018}\natexlab{}.
\newblock \showarticletitle{Social gan: Socially acceptable trajectories with
  generative adversarial networks}. In \bibinfo{booktitle}{\emph{IEEE
  Conference on Computer Vision and Pattern Recognition}}.
  \bibinfo{pages}{2255--2264}.
\newblock


\bibitem[\protect\citeauthoryear{Hastie, Tibshirani, and Friedman}{Hastie
  et~al\mbox{.}}{2009}]%
        {hastie2009elements}
\bibfield{author}{\bibinfo{person}{Trevor Hastie}, \bibinfo{person}{Robert
  Tibshirani}, {and} \bibinfo{person}{Jerome Friedman}.}
  \bibinfo{year}{2009}\natexlab{}.
\newblock \bibinfo{booktitle}{\emph{The elements of statistical learning: data
  mining, inference, and prediction}}.
\newblock \bibinfo{publisher}{Springer Science \& Business Media}.
\newblock


\bibitem[\protect\citeauthoryear{Hochreiter and Schmidhuber}{Hochreiter and
  Schmidhuber}{1997}]%
        {hochreiter1997long}
\bibfield{author}{\bibinfo{person}{Sepp Hochreiter} {and}
  \bibinfo{person}{J{\"u}rgen Schmidhuber}.} \bibinfo{year}{1997}\natexlab{}.
\newblock \showarticletitle{Long short-term memory}.
\newblock \bibinfo{journal}{\emph{Neural computation}} \bibinfo{volume}{9},
  \bibinfo{number}{8} (\bibinfo{year}{1997}), \bibinfo{pages}{1735--1780}.
\newblock


\bibitem[\protect\citeauthoryear{Inc}{Inc}{2021}]%
        {cuebiq2021}
\bibfield{author}{\bibinfo{person}{Cuebiq Inc}.}
  \bibinfo{year}{2021}\natexlab{}.
\newblock \bibinfo{booktitle}{\emph{Data for Good - Cuebiq}}.
\newblock
\urldef\tempurl%
\url{https://www.cuebiq.com/about/data-for-good/}
\showURL{%
\tempurl}


\bibitem[\protect\citeauthoryear{Khanna}{Khanna}{1990}]%
        {khanna1990foundations}
\bibfield{author}{\bibinfo{person}{Tarun Khanna}.}
  \bibinfo{year}{1990}\natexlab{}.
\newblock \bibinfo{booktitle}{\emph{Foundations of neural networks}}.
\newblock \bibinfo{publisher}{Addison-Wesley Longman Publishing Co., Inc.}
\newblock


\bibitem[\protect\citeauthoryear{Li, Li, Gunopulos, and Guibas}{Li
  et~al\mbox{.}}{2016}]%
        {li2016knowledge}
\bibfield{author}{\bibinfo{person}{Yang Li}, \bibinfo{person}{Yangyan Li},
  \bibinfo{person}{Dimitrios Gunopulos}, {and} \bibinfo{person}{Leonidas
  Guibas}.} \bibinfo{year}{2016}\natexlab{}.
\newblock \showarticletitle{Knowledge-based trajectory completion from sparse
  GPS samples}. In \bibinfo{booktitle}{\emph{Proceedings of the 24th ACM
  SIGSPATIAL International Conference on Advances in Geographic Information
  Systems}}. \bibinfo{pages}{1--10}.
\newblock


\bibitem[\protect\citeauthoryear{Liang, Wang, Li, Chen, Li, and Lei}{Liang
  et~al\mbox{.}}{2016}]%
        {liang2016online}
\bibfield{author}{\bibinfo{person}{Biwei Liang}, \bibinfo{person}{Tengjiao
  Wang}, \bibinfo{person}{Shun Li}, \bibinfo{person}{Wei Chen},
  \bibinfo{person}{Hongyan Li}, {and} \bibinfo{person}{Kai Lei}.}
  \bibinfo{year}{2016}\natexlab{}.
\newblock \showarticletitle{Online learning for accurate real-time map
  matching}. In \bibinfo{booktitle}{\emph{PAKDD}}. Springer,
  \bibinfo{pages}{67--78}.
\newblock


\bibitem[\protect\citeauthoryear{Liu, Yu, Zheng, Zhan, and Yue}{Liu
  et~al\mbox{.}}{2019}]%
        {liu2019naomi}
\bibfield{author}{\bibinfo{person}{Yukai Liu}, \bibinfo{person}{Rose Yu},
  \bibinfo{person}{Stephan Zheng}, \bibinfo{person}{Eric Zhan}, {and}
  \bibinfo{person}{Yisong Yue}.} \bibinfo{year}{2019}\natexlab{}.
\newblock \showarticletitle{NAOMI: Non-autoregressive multiresolution sequence
  imputation}. In \bibinfo{booktitle}{\emph{Advances in Neural Information
  Processing Systems}}. \bibinfo{pages}{11238--11248}.
\newblock


\bibitem[\protect\citeauthoryear{Lou, Zhang, Zheng, Xie, Wang, and Huang}{Lou
  et~al\mbox{.}}{2009}]%
        {lou2009map}
\bibfield{author}{\bibinfo{person}{Yin Lou}, \bibinfo{person}{Chengyang Zhang},
  \bibinfo{person}{Yu Zheng}, \bibinfo{person}{Xing Xie}, \bibinfo{person}{Wei
  Wang}, {and} \bibinfo{person}{Yan Huang}.} \bibinfo{year}{2009}\natexlab{}.
\newblock \showarticletitle{Map-matching for low-sampling-rate GPS
  trajectories}. In \bibinfo{booktitle}{\emph{Proceedings of the 17th ACM
  SIGSPATIAL international conference on advances in geographic information
  systems}}. \bibinfo{pages}{352--361}.
\newblock


\bibitem[\protect\citeauthoryear{Luo, Cai, Zhang, Xu, et~al\mbox{.}}{Luo
  et~al\mbox{.}}{2018}]%
        {luo2018multivariate}
\bibfield{author}{\bibinfo{person}{Yonghong Luo}, \bibinfo{person}{Xiangrui
  Cai}, \bibinfo{person}{Ying Zhang}, \bibinfo{person}{Jun Xu},
  {et~al\mbox{.}}} \bibinfo{year}{2018}\natexlab{}.
\newblock \showarticletitle{Multivariate time series imputation with generative
  adversarial networks}. In \bibinfo{booktitle}{\emph{Advances in Neural
  Information Processing Systems}}. \bibinfo{pages}{1596--1607}.
\newblock


\bibitem[\protect\citeauthoryear{Luo, Zhang, Cai, and Yuan}{Luo
  et~al\mbox{.}}{2019}]%
        {luo2019e2gan}
\bibfield{author}{\bibinfo{person}{Yonghong Luo}, \bibinfo{person}{Ying Zhang},
  \bibinfo{person}{Xiangrui Cai}, {and} \bibinfo{person}{Xiaojie Yuan}.}
  \bibinfo{year}{2019}\natexlab{}.
\newblock \showarticletitle{E2GAN: End-to-End Generative Adversarial Network
  for Multivariate Time Series Imputation}. In \bibinfo{booktitle}{\emph{28th
  International Joint Conference on Artificial Intelligence}}. AAAI Press,
  \bibinfo{pages}{3094--3100}.
\newblock


\bibitem[\protect\citeauthoryear{Monreale, Pinelli, Trasarti, and
  Giannotti}{Monreale et~al\mbox{.}}{2009}]%
        {monreale2009wherenext}
\bibfield{author}{\bibinfo{person}{Anna Monreale}, \bibinfo{person}{Fabio
  Pinelli}, \bibinfo{person}{Roberto Trasarti}, {and} \bibinfo{person}{Fosca
  Giannotti}.} \bibinfo{year}{2009}\natexlab{}.
\newblock \showarticletitle{Wherenext: a location predictor on trajectory
  pattern mining}. In \bibinfo{booktitle}{\emph{Proceedings of the 15th ACM
  SIGKDD international conference on Knowledge discovery and data mining}}.
  \bibinfo{pages}{637--646}.
\newblock


\bibitem[\protect\citeauthoryear{Naghizade, Chan, Ren, and Tomko}{Naghizade
  et~al\mbox{.}}{2018}]%
        {naghizade2018contextual}
\bibfield{author}{\bibinfo{person}{Elham Naghizade}, \bibinfo{person}{Jeffrey
  Chan}, \bibinfo{person}{Yongli Ren}, {and} \bibinfo{person}{Martin Tomko}.}
  \bibinfo{year}{2018}\natexlab{}.
\newblock \showarticletitle{Contextual Location Imputation for Confined WiFi
  Trajectories}. In \bibinfo{booktitle}{\emph{Pacific-Asia Conference on
  Knowledge Discovery and Data Mining}}. Springer, \bibinfo{pages}{444--457}.
\newblock


\bibitem[\protect\citeauthoryear{Naghizade, Kulik, Tanin, and Bailey}{Naghizade
  et~al\mbox{.}}{2020}]%
        {naghizade2020privacy}
\bibfield{author}{\bibinfo{person}{Elham Naghizade}, \bibinfo{person}{Lars
  Kulik}, \bibinfo{person}{Egemen Tanin}, {and} \bibinfo{person}{James
  Bailey}.} \bibinfo{year}{2020}\natexlab{}.
\newblock \showarticletitle{Privacy-and context-aware release of trajectory
  data}.
\newblock \bibinfo{journal}{\emph{ACM Transactions on Spatial Algorithms and
  Systems (TSAS)}} \bibinfo{volume}{6}, \bibinfo{number}{1}
  (\bibinfo{year}{2020}), \bibinfo{pages}{1--25}.
\newblock


\bibitem[\protect\citeauthoryear{Qi, Qin, Wu, and Yang}{Qi
  et~al\mbox{.}}{2020}]%
        {qi2020imitative}
\bibfield{author}{\bibinfo{person}{Mengshi Qi}, \bibinfo{person}{Jie Qin},
  \bibinfo{person}{Yu Wu}, {and} \bibinfo{person}{Yi Yang}.}
  \bibinfo{year}{2020}\natexlab{}.
\newblock \showarticletitle{Imitative Non-Autoregressive Modeling for
  Trajectory Forecasting and Imputation}. In
  \bibinfo{booktitle}{\emph{IEEE/CVF}}. \bibinfo{pages}{12736--12745}.
\newblock


\bibitem[\protect\citeauthoryear{Sadri, Salim, Ren, Shao, Krumm, and
  Mascolo}{Sadri et~al\mbox{.}}{2018}]%
        {sadri2018will}
\bibfield{author}{\bibinfo{person}{Amin Sadri}, \bibinfo{person}{Flora~D
  Salim}, \bibinfo{person}{Yongli Ren}, \bibinfo{person}{Wei Shao},
  \bibinfo{person}{John~C Krumm}, {and} \bibinfo{person}{Cecilia Mascolo}.}
  \bibinfo{year}{2018}\natexlab{}.
\newblock \showarticletitle{What will you do for the rest of the day? an
  approach to continuous trajectory prediction}.
\newblock \bibinfo{journal}{\emph{Proc. of the ACM on Interactive, Mobile,
  Wearable and Ubiquitous Technologies}} \bibinfo{volume}{2},
  \bibinfo{number}{4} (\bibinfo{year}{2018}), \bibinfo{pages}{1--26}.
\newblock


\bibitem[\protect\citeauthoryear{Sundermeyer, Schl{\"u}ter, and
  Ney}{Sundermeyer et~al\mbox{.}}{2012}]%
        {sundermeyer2012lstm}
\bibfield{author}{\bibinfo{person}{Martin Sundermeyer}, \bibinfo{person}{Ralf
  Schl{\"u}ter}, {and} \bibinfo{person}{Hermann Ney}.}
  \bibinfo{year}{2012}\natexlab{}.
\newblock \showarticletitle{LSTM neural networks for language modeling}. In
  \bibinfo{booktitle}{\emph{Thirteenth annual conference of the international
  speech communication association}}.
\newblock


\bibitem[\protect\citeauthoryear{Teixeira, Viana, Almeida, and Alvim}{Teixeira
  et~al\mbox{.}}{2021}]%
        {teixeira2021impact}
\bibfield{author}{\bibinfo{person}{Douglas Do~Couto Teixeira},
  \bibinfo{person}{Aline~Carneiro Viana}, \bibinfo{person}{Jussara~M Almeida},
  {and} \bibinfo{person}{Mrio~S Alvim}.} \bibinfo{year}{2021}\natexlab{}.
\newblock \showarticletitle{The impact of stationarity, regularity, and context
  on the predictability of individual human mobility}.
\newblock \bibinfo{journal}{\emph{ACM Transactions on Spatial Algorithms and
  Systems}} \bibinfo{volume}{7}, \bibinfo{number}{4} (\bibinfo{year}{2021}),
  \bibinfo{pages}{1--24}.
\newblock


\bibitem[\protect\citeauthoryear{Vaswani, Shazeer, Parmar, Uszkoreit, Jones,
  Gomez, Kaiser, and Polosukhin}{Vaswani et~al\mbox{.}}{2017}]%
        {vaswani2017attention}
\bibfield{author}{\bibinfo{person}{Ashish Vaswani}, \bibinfo{person}{Noam
  Shazeer}, \bibinfo{person}{Niki Parmar}, \bibinfo{person}{Jakob Uszkoreit},
  \bibinfo{person}{Llion Jones}, \bibinfo{person}{Aidan~N Gomez},
  \bibinfo{person}{{\L}ukasz Kaiser}, {and} \bibinfo{person}{Illia
  Polosukhin}.} \bibinfo{year}{2017}\natexlab{}.
\newblock \showarticletitle{Attention is all you need}. In
  \bibinfo{booktitle}{\emph{Advances in neural information processing
  systems}}. \bibinfo{pages}{5998--6008}.
\newblock


\bibitem[\protect\citeauthoryear{Veli{\v{c}}kovi{\'c}, Cucurull, Casanova,
  Romero, Lio, and Bengio}{Veli{\v{c}}kovi{\'c} et~al\mbox{.}}{2017}]%
        {velivckovic2017graph}
\bibfield{author}{\bibinfo{person}{Petar Veli{\v{c}}kovi{\'c}},
  \bibinfo{person}{Guillem Cucurull}, \bibinfo{person}{Arantxa Casanova},
  \bibinfo{person}{Adriana Romero}, \bibinfo{person}{Pietro Lio}, {and}
  \bibinfo{person}{Yoshua Bengio}.} \bibinfo{year}{2017}\natexlab{}.
\newblock \showarticletitle{Graph attention networks}.
\newblock \bibinfo{journal}{\emph{arXiv preprint arXiv:1710.10903}}
  (\bibinfo{year}{2017}).
\newblock


\bibitem[\protect\citeauthoryear{Wang, Shen, Ouyang, and Cheng}{Wang
  et~al\mbox{.}}{2018}]%
        {wang2018exploiting}
\bibfield{author}{\bibinfo{person}{Hao Wang}, \bibinfo{person}{Huawei Shen},
  \bibinfo{person}{Wentao Ouyang}, {and} \bibinfo{person}{Xueqi Cheng}.}
  \bibinfo{year}{2018}\natexlab{}.
\newblock \showarticletitle{Exploiting POI-Specific Geographical Influence for
  Point-of-Interest Recommendation.}. In \bibinfo{booktitle}{\emph{IJCAI}}.
  \bibinfo{pages}{3877--3883}.
\newblock


\bibitem[\protect\citeauthoryear{Wang, Salim, Ren, and Koniusz}{Wang
  et~al\mbox{.}}{2020}]%
        {wang2020relation}
\bibfield{author}{\bibinfo{person}{Xianjing Wang}, \bibinfo{person}{Flora~D
  Salim}, \bibinfo{person}{Yongli Ren}, {and} \bibinfo{person}{Piotr Koniusz}.}
  \bibinfo{year}{2020}\natexlab{}.
\newblock \showarticletitle{Relation Embedding for Personalised
  Translation-Based POI Recommendation}. In
  \bibinfo{booktitle}{\emph{Pacific-Asia Conference on Knowledge Discovery and
  Data Mining}}. Springer, \bibinfo{pages}{53--64}.
\newblock


\bibitem[\protect\citeauthoryear{Yin, Shah, Wang, and Zimmermann}{Yin
  et~al\mbox{.}}{2018}]%
        {yin2018feature}
\bibfield{author}{\bibinfo{person}{Yifang Yin}, \bibinfo{person}{Rajiv~Ratn
  Shah}, \bibinfo{person}{Guanfeng Wang}, {and} \bibinfo{person}{Roger
  Zimmermann}.} \bibinfo{year}{2018}\natexlab{}.
\newblock \showarticletitle{Feature-based map matching for low-sampling-rate
  GPS trajectories}.
\newblock \bibinfo{journal}{\emph{ACM Transactions on Spatial Algorithms and
  Systems (TSAS)}} \bibinfo{volume}{4}, \bibinfo{number}{2}
  (\bibinfo{year}{2018}), \bibinfo{pages}{1--24}.
\newblock


\bibitem[\protect\citeauthoryear{Yoon, Jordon, and Schaar}{Yoon
  et~al\mbox{.}}{2018a}]%
        {yoon2018gain}
\bibfield{author}{\bibinfo{person}{Jinsung Yoon}, \bibinfo{person}{James
  Jordon}, {and} \bibinfo{person}{Mihaela Schaar}.}
  \bibinfo{year}{2018}\natexlab{a}.
\newblock \showarticletitle{Gain: Missing data imputation using generative
  adversarial nets}. In \bibinfo{booktitle}{\emph{International Conference on
  Machine Learning}}. PMLR, \bibinfo{pages}{5689--5698}.
\newblock


\bibitem[\protect\citeauthoryear{Yoon, Zame, and van~der Schaar}{Yoon
  et~al\mbox{.}}{2018b}]%
        {yoon2018estimating}
\bibfield{author}{\bibinfo{person}{Jinsung Yoon}, \bibinfo{person}{William~R
  Zame}, {and} \bibinfo{person}{Mihaela van~der Schaar}.}
  \bibinfo{year}{2018}\natexlab{b}.
\newblock \showarticletitle{Estimating missing data in temporal data streams
  using multi-directional recurrent neural networks}.
\newblock \bibinfo{journal}{\emph{IEEE Transactions on Biomedical Engineering}}
  \bibinfo{volume}{66}, \bibinfo{number}{5} (\bibinfo{year}{2018}),
  \bibinfo{pages}{1477--1490}.
\newblock


\bibitem[\protect\citeauthoryear{Zheng, Zheng, Xie, and Zhou}{Zheng
  et~al\mbox{.}}{2012}]%
        {zheng2012reducing}
\bibfield{author}{\bibinfo{person}{Kai Zheng}, \bibinfo{person}{Yu Zheng},
  \bibinfo{person}{Xing Xie}, {and} \bibinfo{person}{Xiaofang Zhou}.}
  \bibinfo{year}{2012}\natexlab{}.
\newblock \showarticletitle{Reducing uncertainty of low-sampling-rate
  trajectories}. In \bibinfo{booktitle}{\emph{2012 IEEE 28th international
  conference on data engineering}}. IEEE, \bibinfo{pages}{1144--1155}.
\newblock


\bibitem[\protect\citeauthoryear{Zheng, Zhang, Xie, and Ma}{Zheng
  et~al\mbox{.}}{2009}]%
        {zheng2009mining}
\bibfield{author}{\bibinfo{person}{Yu Zheng}, \bibinfo{person}{Lizhu Zhang},
  \bibinfo{person}{Xing Xie}, {and} \bibinfo{person}{Wei-Ying Ma}.}
  \bibinfo{year}{2009}\natexlab{}.
\newblock \showarticletitle{Mining interesting locations and travel sequences
  from GPS trajectories}. In \bibinfo{booktitle}{\emph{18th International
  conference on World Wide Web}}. \bibinfo{pages}{791--800}.
\newblock


\bibitem[\protect\citeauthoryear{Zhou, Wu, Trajcevski, Khokhar, and Zhang}{Zhou
  et~al\mbox{.}}{2020}]%
        {zhou2020semi}
\bibfield{author}{\bibinfo{person}{Fan Zhou}, \bibinfo{person}{Hantao Wu},
  \bibinfo{person}{Goce Trajcevski}, \bibinfo{person}{Ashfaq Khokhar}, {and}
  \bibinfo{person}{Kunpeng Zhang}.} \bibinfo{year}{2020}\natexlab{}.
\newblock \showarticletitle{Semi-supervised trajectory understanding with poi
  attention for end-to-end trip recommendation}.
\newblock \bibinfo{journal}{\emph{ACM Transactions on Spatial Algorithms and
  Systems (TSAS)}} \bibinfo{volume}{6}, \bibinfo{number}{2}
  (\bibinfo{year}{2020}), \bibinfo{pages}{1--25}.
\newblock


\end{thebibliography}

\end{document}